\documentclass[journal]{IEEEtran}
\usepackage[noadjust]{cite}
\ifCLASSINFOpdf
\else
\fi

\usepackage{bm}
\usepackage{algorithm}  
\usepackage{algorithmicx} 
\usepackage{algpseudocode}  
\usepackage{caption}
\usepackage{graphicx}

\usepackage{amsfonts}
\usepackage{soul}
\usepackage{color}
\usepackage{xcolor}
\usepackage{multirow}
\usepackage{booktabs}
\usepackage{siunitx}
\usepackage{amsmath}
\usepackage[colorlinks=true,citecolor=green]{hyperref}
\usepackage{enumitem}

\usepackage{pdflscape}

\DeclareMathOperator*{\argmin}{arg\,min}

\makeatletter
\newcommand{\algmargin}{\the\ALG@thistlm}
\makeatother
\newlength{\whilewidth}
\algdef{SE}[parWHILE]{parWhile}{EndparWhile}[1]
  {\parbox[t]{\dimexpr\linewidth-\algmargin}{%
     \hangindent\whilewidth\strut\algorithmicwhile\ #1\ \algorithmicdo\strut}}{\algorithmicend\ \algorithmicwhile}%
\algnewcommand{\parState}[1]{\State%
  \parbox[t]{\dimexpr\linewidth-\algmargin}{\strut #1\strut}}

\algdef{SE}[SUBALG]{Indent}{EndIndent}{}{\algorithmicend\ }%
\algtext*{Indent}
\algtext*{EndIndent}

\usepackage{tabularx, makecell, booktabs}

\begin{document}
\title{Towards Personalized Federated Learning}
%A Survey of Personalized Federated Learning: Challenges, Techniques and Future Directions}

\author{Alysa Ziying Tan,
        Han Yu$^{*}$,
        Lizhen Cui$^{*}$,
        and Qiang Yang$^{*}$,~\IEEEmembership{Fellow,~IEEE}% <-this % stops a space
\thanks{Alysa Ziying Tan is with the School of Computer Science and Engineering, Nanyang Technological University, Singapore; Alibaba-NTU Singapore Joint Research Institute, NTU, Singapore; and Alibaba Group, Hangzhou, China.}
\thanks{Han Yu is with the School of Computer Science and Engineering, Nanyang Technological University, Singapore.}% <-this % stops a space
\thanks{Lizhen Cui is with the School of Software, Shandong University (SDU), Jinan, China; and the Joint SDU-NTU Centre for Artificial Intelligence Research (C-FAIR), SDU, Jinan, China.}% <-this % stops a space
\thanks{Qiang Yang is with the Department of Computer Science and Engineering, Hong Kong University of Science and Technology, Hong Kong; and WeBank, Shenzhen, China.} \thanks{$^{*}$Corresponding authors: Han Yu (han.yu@ntu.edu.sg), Lizhen Cui (clz@sdu.edu.cn) and Qiang Yang (qyang@cse.ust.hk)}
}


\markboth{IEEE Transactions on Neural Networks and Learning Systems}%
{Shell \MakeLowercase{\textit{et al.}}: Bare Demo of IEEEtran.cls for IEEE Journals}



\maketitle

\begin{abstract}
In parallel with the rapid adoption of Artificial Intelligence (AI) empowered by advances in AI research, there have been growing awareness and concerns of data privacy. Recent significant developments in the data regulation landscape have prompted a seismic shift in interest towards privacy-preserving AI. This has contributed to the popularity of Federated Learning (FL), the leading paradigm for the training of machine learning models on data silos in a privacy-preserving manner. In this survey, we explore the domain of Personalized FL (PFL) to address the fundamental challenges of FL on heterogeneous data, a universal characteristic inherent in all real-world datasets. We analyze the key motivations for PFL and present a unique taxonomy of PFL techniques categorized according to the key challenges and personalization strategies in PFL. We highlight their key ideas, challenges and opportunities and envision promising future trajectories of research towards new PFL architectural design, realistic PFL benchmarking, and trustworthy PFL approaches.

\end{abstract}


% Enter key words or phrases in alphabetical order, separated by commas. For a list of suggested keywords, send a blank e-mail to keywords@ieee.org or visit http://www.ieee.org/organizations/pubs/ani\_prod/keywrd98.txt

% \fi
\IEEEpeerreviewmaketitle

\section{Introduction}

\IEEEPARstart{T}{he} pervasiveness of edge devices in modern society, such as mobile phones and wearable devices, has led to the rapid growth of private data originating from distributed sources. In this digital age, organizations are using big data and artificial intelligence (AI) to optimize their processes and performance. While the wealth of data offers tremendous opportunities for AI applications, most of these data are highly-sensitive in nature and they exist in the form of isolated islands. This is especially relevant in the healthcare industry where medical data are highly-sensitive and they are often collected and reside across different healthcare institutions \cite{kaissisSecurePrivacypreservingFederated2020,warnat-herresthalSwarmLearningDecentralized2021,shellerFederatedLearningMedicine2020,dayan2021federated}. Such circumstances pose huge challenges for AI adoption as data privacy issues are not well addressed by conventional AI approaches. With the recent introduction of data privacy preservation laws such as the General Data Protection Regulation (GDPR) \cite{voigtEUGeneralData2017}, there is an increasing demand for privacy-preserving AI \cite{cheng2020federated} in order to meet regulatory compliance.

In view of these data privacy challenges, Federated Learning (FL) \cite{yangFederatedMachineLearning2019e, kairouzAdvancesOpenProblems2019a} has seen growing popularity in recent years. FL is a learning paradigm that enables collaborative training of machine learning models involving multiple data silos in a privacy-preserving manner. 
The prevailing FL setting assumes a federation of data owners (a.k.a. clients), which may be as small as individual mobile devices to as large as entire organizations, that collaboratively train a model under the orchestration of a central parameter server (a.k.a. the FL server) \cite{yangFederatedMachineLearning2019e, kairouzAdvancesOpenProblems2019a}. The training data are stored locally and are not directly shared during the training process. Most of the existing FL training approaches are derived from the Federated Averaging (FedAvg) algorithm introduced in \cite{mcmahanCommunicationEfficientLearningDeep2017}. The goal is to train a global model that performs well on most FL clients.

\begin{figure*}[t!]
\centering
\includegraphics[width=1\linewidth]{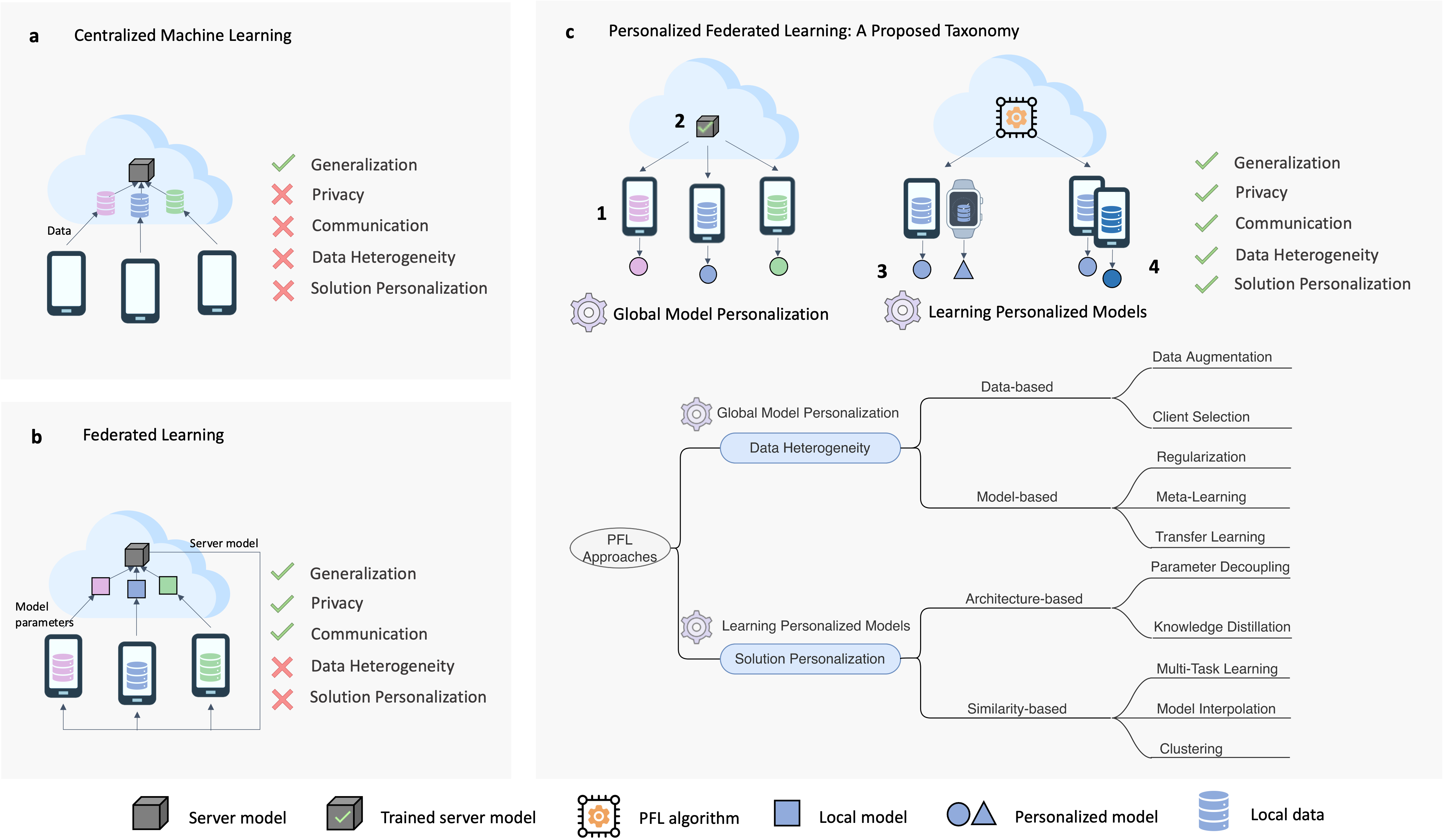}
\caption{Concept, Motivations \& Proposed Taxonomy for Personalized Federated learning. \textbf{a}. Centralized machine learning (CML) which pools data together to train a central ML model. \textbf{b}. Federated learning (FL) which trains a global model under the orchestration of a central parameter server. Data resides in different data silos. \textbf{c}. Personalized federated learning (PFL) which addresses the limitations of FL through global model personalization and personalized models learning.  \textbf{1--4} Four categories of PFL approaches: \textbf{1}) data-based, \textbf{2}) model-based \textbf{3}) architecture-based, \textbf{4}) similarity-based.}
\label{ml_fl_pfl}
\end{figure*}

\subsection{Categorization of Federated Learning}
% HFL vs VFL vs FTL
FL can be categorized into horizontal FL (HFL), vertical FL (VFL) and federated transfer learning (FTL), according to how data are distributed in terms of feature and sample spaces among participating entities \cite{yangFederatedMachineLearning2019e}. HFL refers to scenarios whereby participants share the same feature space but have different data samples. It is the most commonly adopted FL setting popularized by Google, which applied HFL to train language models in mobile devices \cite{mcmahanCommunicationEfficientLearningDeep2017}. In VFL, participants have overlapping data samples, but differ in the feature space. A typical application scenario would involve the collaboration of multiple organizations from different industry sectors (e.g., a bank and an e-commerce company) which have different data features but may have a large number of shared users. FTL is applicable when participants have little overlap in both the feature space and the sample space. For example, organizations from different industry sectors serving markets in different regions can leverage FTL to collaboratively build models.
Existing PFL works mainly focus on the HFL setting which makes up the majority of the FL application scenarios \cite{kairouzAdvancesOpenProblems2019a}. The HFL setting is the focus of this paper. For brevity, we use the terms HFL and FL interchangeably in the rest of this survey.

\subsection{Motivations for Personalized Federated Learning}

Fig. \ref{ml_fl_pfl} illustrates the key concepts and motivations for centralized machine learning (CML) \cite{drainakis2020federated}, FL and PFL. We consider a cloud-based CML setting where data are pooled together in the cloud server to train an ML model. In this setting, the CML model achieves good generalization from the rich amount of data. However, CML faces bandwidth and latency challenges due to the sheer amount of data transferred to the cloud. It also does not preserve data privacy or not personalize well.

The FL setting assumes a federation of distributed clients, each with its own private local dataset. As these clients face data scarcity that limit their capacities to train effective local models, they are motivated to join the FL process to obtain a better performing model. FL enables collaborative model training on data silos in a privacy-preserving manner, which sets it apart from the CML setting. Additionally, FL is communication-efficient as it only transfers model parameters which are a fraction in size compared to transferring raw data. By considering privacy and communication constraints, FL is applicable to support a wide range of application scenarios such as Internet of Things (IoT), that entails privacy, connectivity, bandwidth and latency challenges in varying edge computing environments \cite{lim2020federated}.

However, the general FL approach faces several fundamental challenges: (i) poor convergence on highly heterogeneous data, and (ii) lack of solution personalization. These issues deteriorate the performance of the global FL model on individual clients in the presence of heterogeneous local data distributions, and may even disincentivize affected clients from joining the FL process.
% The key differences in motivations for ML, FL and PFL are summarized in Table \ref{general-scope}.
Compared to traditional FL, PFL research seeks to address these two challenges.

\subsubsection{Poor Convergence on Heterogeneous Data} 
\begin{figure}[ht]
\centering
\includegraphics[width=1\linewidth]{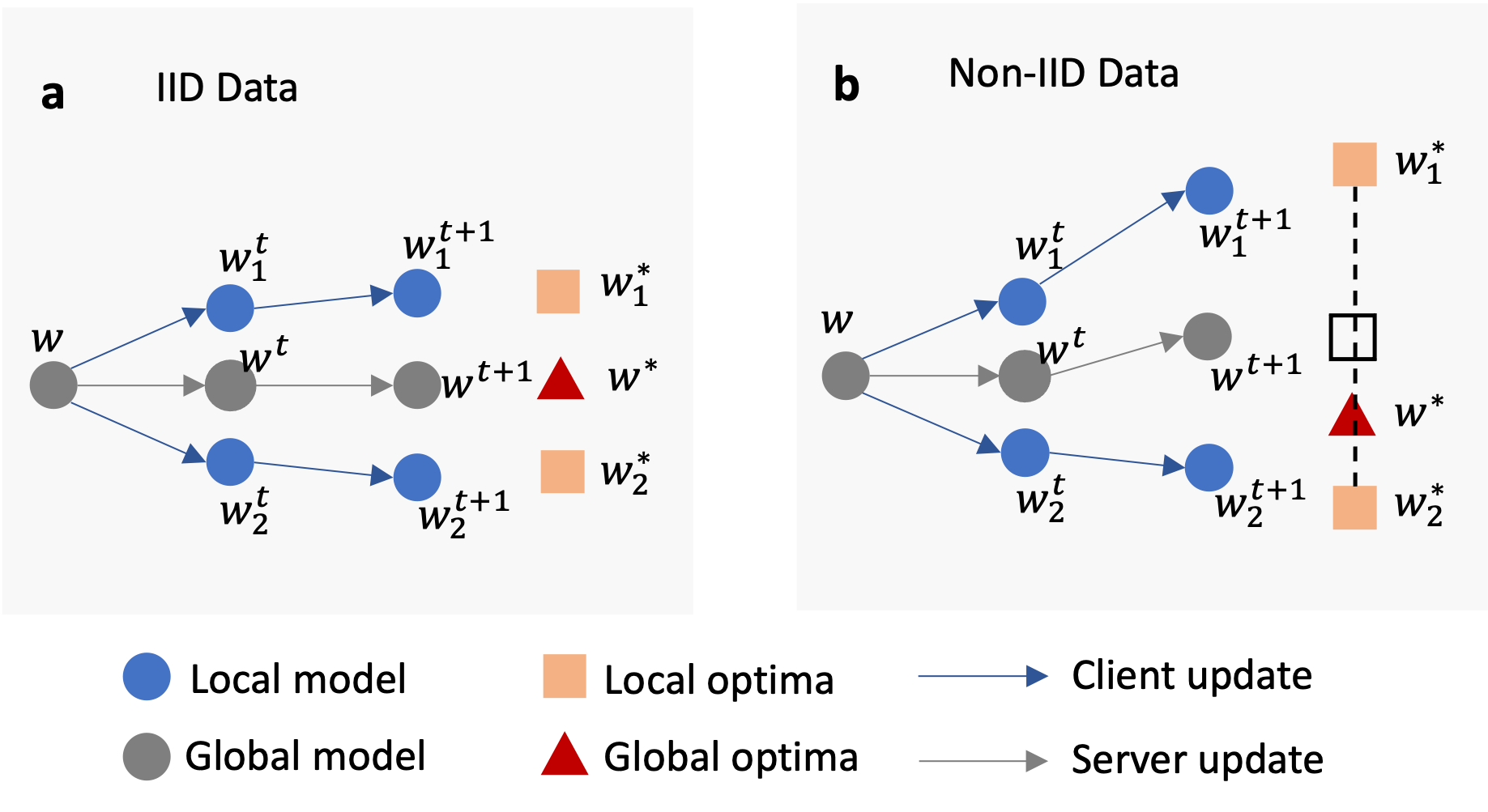}
\caption{Illustration of client drift in FedAvg for 2 clients with 2 local steps. \textbf{a} IID data setting. \textbf{b} Non-IID data setting.}
\label{client-drift}
\end{figure}

When learning on non-independent and identically distributed (non-IID) data, the accuracy of FedAvg is significantly reduced. This performance degradation is attributed to the phenomenon of client drift \cite{karimireddySCAFFOLDStochasticControlled}, as a result of the rounds of local training and synchronization on local data distributions that are non-IID. Fig. \ref{client-drift} illustrates the effect of client drift on IID and non-IID data. In FedAvg, the server updates move toward the average of client optima. When data are IID, the averaged model is close to the global optimum $w^*$ as it is equidistant to both local optima $w_1^*$ and $w_2^*$. However, when data are non-IID, the global optimum $w^*$ is not equidistant to the local optima. In this illustration, $w^*$ is closer to $w_2^*$. The averaged model $w^{t+1}$ will therefore be far from the global optimum $w^*$, and the global model does not converge to its true global optimum. As the FedAvg algorithm experiences convergence issues on non-IID data, careful tuning of hyperparameters (e.g., learning rate decay) is required to improve learning stability \cite{liConvergenceFedAvgNonIID2020}.

\subsubsection{Lack of Solution Personalization}
In the vanilla FL setting, a single globally-shared model is trained to fit the ``average client''. As a result, the global model will not generalize well for a local distribution that is very different from the global distribution. Having a single model is often insufficient for practical applications which often face non-IID local datasets. Taking the example of applying FL to develop language models for mobile keyboards, users from different demographics are likely to have divergent usage patterns due to diverse generational, linguistic and cultural nuances. Certain words or emojis are likely be used predominantly by specific groups of users. For such a scenario, a more tailored prediction pattern is needed for each individual user in order for the word suggestions to be meaningful.

\subsection{Contributions}
There are several surveys on the general concepts, methods and applications of FL \cite{yangFederatedMachineLearning2019e,liFederatedLearningChallenges2020d}. Others review FL from the perspectives of privacy \cite{mothukuriSurveySecurityPrivacy2021} and robustness \cite{lyuThreatsFederatedLearning2020a}. Our survey focuses on PFL, which studies the problem of learning personalized models to handle statistical heterogeneity under the FL setting. There is a shortage of a comprehensive survey on PFL that provides a systematic perspective on this important topic for new researchers.
In this paper, we bridge this gap in the current FL literature. Our main contributions are summarized as follows:
\begin{itemize}
     \item We provide a succinct overview of FL and its categorization. A detailed analysis of the key motivations for PFL in the current FL settings is also included.
     \item We identify personalization strategies to address key FL challenges, and offer a unique data-based, model-based, architecture-based and similarity-based perspective for guiding the review of the PFL literature. Based on this perspective, we propose a hierarchical taxonomy to present existing works on PFL, highlighting the challenges they face, their main ideas and assumptions they made which could introduce potential limitations.
     \item We discuss commonly adopted public datasets and evaluation metrics in the current literature for PFL benchmarking, and offer suggestions on enhancing PFL experimental evaluation techniques.
     \item We envision promising future trajectories of research towards new architectural design, realistic benchmarking, and trustworthy approaches towards building personalized federated learning systems.
 \end{itemize}
\section{Strategies for Personalized Federated Learning}
In this section, we provide an overview of the PFL strategies which are the basis for our systematic and comprehensive review of existing PFL approaches. We organize the literature around the proposed taxonomy (Fig. \ref{ml_fl_pfl}c) that divides PFL methods according to the key challenges and personalization strategies involved. \bigskip

\noindent\textit{Strategy I: Global Model Personalization}
\newline

The first strategy addresses the performance issues in training a globally-shared FL model on heterogeneous data. When learning on non-IID data, the accuracy of FedAvg-based approaches is significantly reduced due to client drift. Under global model personalization, the PFL setup closely follows the general FL training procedure where a single global FL model is trained. The trained global FL model is then personalized for each FL client through a local adaptation step that involves additional training on each local dataset. This two-step ``FL training + local adaptation'' approach is commonly regarded as an FL personalization strategy by the FL community  \cite{kairouzAdvancesOpenProblems2019a, mansourThreeApproachesPersonalization2020a}. As personalization performance directly depends on the generalization performance of the global model, many PFL approaches aim to improve the performance of the global model under data heterogeneity in order to improve the performance of subsequent personalization on local data. Personalization techniques for this category are classified into data-based and model-based approaches. Data-based approaches aim to mitigate the client drift problem by reducing the statistical heterogeneity among the clients' datasets, while model-based approaches aim to learn a strong global model for future personalization on individual clients or improve the adaptation performance of the local model.\bigskip

\noindent\textit{Strategy II: Learning Personalized Models}
\newline
The second strategy addresses the challenge of solution personalization. In contrast to the global model personalization strategy which trains a single global model, approaches in this category train individual personalized FL models. The goal is to build personalized models by modifying the FL model aggregation process. This is achieved through applying different learning paradigms in the FL setting. Personalization techniques are classified into architecture-based and similarity-based approaches. Architecture-based approaches aim to provide a personalized model architecture tailored to each client, while similarity-based approaches aim to leverage client relationships to improve personalized model performance where similar personalized models are built for related clients.

In personalized FL model training, the optimization objective is formulated differently from the vanilla FL setting, as an individual personalized model is learned for each client. Here, we provide formulations of the optimization objectives under the FL setting and the local learning setting in order to highlight the positioning of PFL approaches. 
The standard FL objective is given as 
\begin{equation}
\label{eq:std_loss}
    \min_{w \in \mathbb{R}^{d}}F\left(w\right) := \frac{1}{C}\sum_{c=1}^{C}f_c\left(w\right),
\end{equation}
where $C$ is the number of participating clients, $w \in \mathbb{R}^d$ encodes the parameters of the
global model and
\begin{equation}
\label{eq:local_loss2}
    f_c\left(w\right) := \mathbb{E}_{(x, y)\sim D_c}\left[f_c\left(w;x,y\right)\right]
\end{equation}
represents the expected loss over the data distribution $D_c$ of client $c$. The prevailing FL formulation minimizes the aggregation of local functions and entails a common output for all clients using the global model without any personalization.  In the presence of data heterogeneity (i.e., the underlying data distributions across the clients are not identical), simply minimizing the average local loss with no personalization will result in poor performance. 

At the opposite end of the spectrum, we consider a local learning setting where each client $c$ trains its own model $\theta_c$ locally without any communication with other clients. The objective is given as 
\begin{equation}
\label{eq:full_per_loss}
    \min_{\theta_1, ..., \theta_c \in \mathbb{R}^{d}}F\left(\theta\right) := \frac{1}{C}\sum_{c=1}^{C}f_c\left(\theta_c\right),
\end{equation}
where $\theta_c \in \mathbb{R}^d$ encodes the parameters of the local model of client $c$. In this setting, the resulting models may not achieve good generalization performance as the number of training examples that the local models are exposed to are limited. Stronger generalization guarantees can be obtained with more collaboration amongst clients to exploit the pool of knowledge for model training. 

Comparing the formulations of the standard FL and local learning settings, standard FL facilitates collaboration and knowledge sharing amongst clients but does not entail personalized outputs as it relies on a shared global model for client inference. On the other hand, local learning entails a fully personalized model for each client, but fails to leverage potential performance gains from inter-client collaboration. Given the need to achieve a balance between generalization and personalization performance, PFL approaches fall between the standard FL setting and the local learning setting.

\section{Strategy I: Global Model Personalization}
\begin{figure*}[t!]
\centering
\includegraphics[width=1\linewidth]{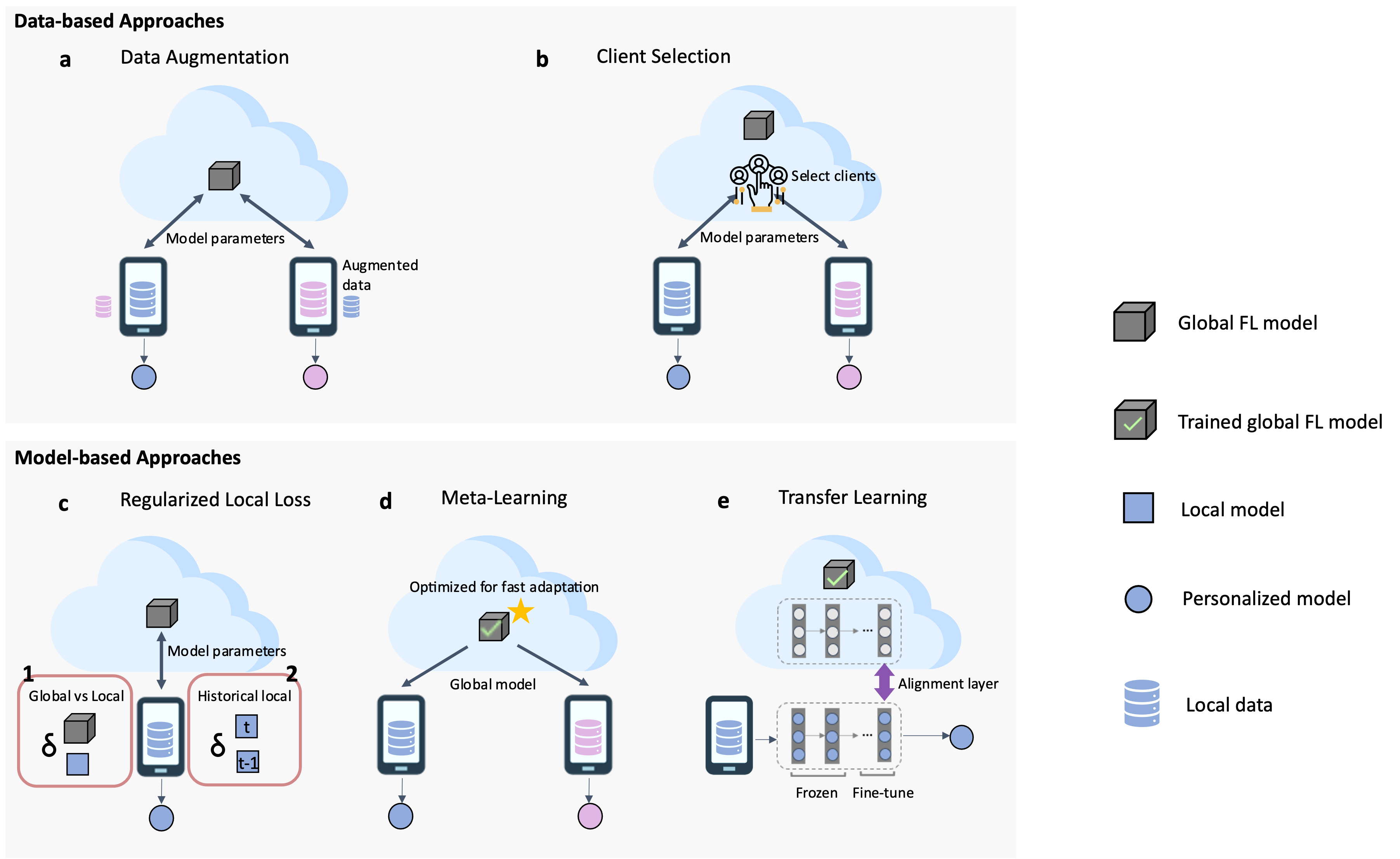}
\caption{The setup \& configurations of approaches that fall under \textit{Strategy I: Global Model Personalization}. \textbf{a--b} Data-based approaches: (\textbf{a}) data augmentation, (\textbf{b}) client selection. \textbf{c--e} Model-based approaches: (\textbf{c}) regularized local loss; regularization can be performed \textbf{1}) between global and local models, \textbf{2}) between historical local model snapshots, (\textbf{d}) meta-learning, (\textbf{e}) transfer learning.}
\label{data_model}
\end{figure*}

In this section, we survey PFL approaches following the global model personalization strategy. The main setup and configurations for these approaches are illustrated in Fig. \ref{data_model}. Based on our proposed taxonomy, they are divided into \textit{Data-based Approaches} and \textit{Model-based Approaches} as follows.

\subsection{Data-based Approaches}

Motivated by the client drift problem arising from federated training on heterogeneous data, data-based approaches aim to reduce the statistical heterogeneity of client data distributions. This helps to improve the generalization performance of the global FL model. \bigskip

\noindent\textit{Data Augmentation}
\newline
As the IID property of training data is a fundamental assumption in statistical learning theory, data augmentation methods to enhance the statistical homogeneity of the data have been extensively studied in the field of machine learning. Over-sampling techniques involving synthetic data generation (e.g., SMOTE \cite{chawlaSMOTESyntheticMinority2002} and ADASYN \cite{haiboheADASYNAdaptiveSynthetic2008}), and under-sampling techniques (e.g., Tomek links \cite{kubatAddressingCurseImbalanced1997}) have been proposed to reduce data imbalance. These techniques, however, cannot be directly applied under the FL setting, where data residing at the clients in the federation are distributed and private.

Data augmentation in FL (Fig. \ref{data_model}a) is highly challenging as it often requires some form of data sharing or relies on the availability of a proxy dataset that is representative of the overall data distribution. In \cite{zhaoFederatedLearningNonIID2018a}, the authors proposed a data sharing strategy that distributes a small amount of global data balanced by classes to each client. Their experiments show that there is potential for significant accuracy gains ($\sim$30\%) with the addition of a small amount of data. In \cite{jeongCommunicationEfficientOnDeviceMachine2018b}, the authors proposed FAug, a federated augmentation approach that involves training a Generative Adversarial Network (GAN) model in the FL server. Some data samples of the minority classes are uploaded to the server to train the GAN model. The trained GAN model is then distributed to each client to generate additional data to augment its local data to produce an IID dataset. In \cite{duanSelfBalancingFederatedLearning2021}, the authors proposed Astraea, a self-balancing FL framework to handle class imbalance by using Z-score based data augmentation and down-sampling of local data. The FL server requires statistical information about clients' local data distributions (e.g., class sizes, mean and standard deviation values). In \cite{wuFedHomeCloudEdgeBased2020}, the authors proposed the FedHome algorithm that trains a Generative Convolutional Autoencoder (GCAE) model using FL. At the end of the FL procedure, each client performs further personalization on a locally augmented class-balanced dataset. This dataset is generated by executing the SMOTE algorithm on the low dimensional features of the encoder network based on the local data. \bigskip

\noindent\textit{Client Selection}
\newline
Another line of work focuses on designing FL client selection mechanisms to enable sampling from a more homogeneous data distribution, with the aim of improving model generalization performance (Fig. \ref{data_model}b). In \cite{wangOptimizingFederatedLearning2020}, the authors proposed FAVOR which selects a subset of participating clients for each training round in order to mitigate the bias introduced by non-IID data. A deep Q-learning formulation for client selection was designed with the objective of maximizing accuracy, while minimizing the number of communication rounds. In a similar approach, a client selection algorithm based on the Multi-Armed Bandit formulation was proposed in \cite{yangFederatedLearningClass2020} to select the subset of clients with minimal class imbalance. The local class distributions are estimated by comparing the similarity between the local gradient updates submitted to the FL server with the gradients inferred from a balanced proxy dataset residing on the server. 

Recently, there is an emerging line of work that focuses on developing client selection strategies to tackle data and resource heterogeneity challenges that are prevalent in edge computing applications. For cross-device FL, there is often significant variability in hardware capabilities in terms of computation and communication capacities. Heterogeneity also exists in data, whereby the quantity and distribution of data differ among clients. Such diversity exacerbates challenges such as communication costs, stragglers and model accuracy. In \cite{chaiTiFLTierbasedFederated2020}, the authors proposed a tier-based FL system (TiFL) that groups clients into tiers based on training performance. The algorithm adaptively selects participating clients from the same tier for each training round by optimizing both accuracy and training time. This helps alleviate the performance issues caused by data and resource heterogeneity. In \cite{liFedSAENovelSelfAdaptive2021}, the authors proposed FedSAE, a self-adaptive FL system that adaptively selects clients with larger local training loss in each training round to accelerate the convergence of the global model. A prediction mechanism of the affordable workload of each client is also proposed to enable the dynamic adjustment of the number of local training epochs for each client in order to improve device reliability.

\subsection{Model-based Approaches}
%\newline
Although data-based approaches improve the convergence of the global FL model by mitigating the client drift problem, they generally need to modify the local data distributions. This may result in loss of valuable information associated with the inherent diversity of client behaviors. Such information can be useful for personalizing the global model for each client. In this section, we cover model-based global model personalization FL approaches. The objective is either to learn a strong global FL model for future personalization on each individual client, or to improve the adaptation performance of the local model. \bigskip

\noindent\textit{Regularized Local Loss}
\newline
Model regularization is a common strategy for preventing overfitting and improving convergence when training machine learning models. In FL, regularization techniques can be applied to limit the impact of local updates. This improves convergence stability and the generalization of the global model, which in turn, can be used to produce better personalized models. Instead of just minimizing the local function $f_c(\theta)$, each client $c$ minimizes the following objective:
\begin{equation}
\label{eq:reg_loss}
    \min_{\theta \in \mathbb{R}^{d}}h_c\left(\theta; w\right) := f_c\left(\theta\right) + l_{reg}\left(\theta; w\right),
\end{equation}
where $l_{reg}\left(\theta; w\right)$ is the regularization loss, which is generally formulated as a function of the global model $w$ and the local model $\theta_c$ of client $c$. Regularization can be applied in the following ways as illustrated in Fig. \ref{data_model}c:
%\newline

\subsubsection{Between Global and Local Models} Several works implement regularization between the global and local models to tackle the client drift problem that is prevalent in FL due to statistical data heterogeneity. FedProx \cite{liFederatedOptimizationHeterogeneous2020b} introduced a proximal term %$\frac{\mu}{2}\lVert\theta_c-w\rVert^2$ 
to the local sub-problem which considers the dissimilarity between the global FL model and local models to adjust the impact of local updates. Along with model dissimilarity, FedCL \cite{yaoContinualLocalTraining2020} further considers parameter importance in the regularized local loss function using Elastic Weight Consolidation (EWC) \cite{kirkpatrickOvercomingCatastrophicForgetting2017} from the field of continual learning. The importance of the weights %$\Omega$ 
to the global model is estimated on a proxy dataset in the FL server. They are then transferred to the clients where penalization steps are carried out to prevent important parameters of the global model from being changed when adapting the global model to clients' local data. Doing so alleviates the weight divergence between the local and global models, while preserving the knowledge of the global model to improve generalization.
Recently, SCAFFOLD \cite{karimireddySCAFFOLDStochasticControlled} uses variance reduction to alleviate the effect of client drifting that causes weight divergence between the local and global models. The update directions of the global ($v$) and local ($v_c$) models are estimated. The difference, $(v-v_c)$, is added as a component of the local loss function to correct local updates.

\subsubsection{Between Historical Local Model Snapshots}
Recently, a contrastive learning-based FL -- MOON \cite{liModelContrastiveFederatedLearning2021} has been proposed. The goal of MOON is to reduce the distance between the representations learned by the local models and the global model (i.e., to alleviate weight divergence), and increase the distance between the representations learned between a given local model and its previous local model (i.e., to speed up convergence). This emerging approach enables each client to learn a representation close to the global model to minimize local model divergence. It also speeds up learning by encouraging the local model to improve from its previous version. \bigskip

% \subsection{Personalization Algorithms}
\noindent\textit{Meta-learning}
\newline
Commonly known as ``learning to learn'', meta-learning aims to improve the learning algorithm through exposure to a variety of tasks (i.e., datasets) \cite{hospedalesMetaLearningNeuralNetworks2021}. This enables the model to learn a new task quickly and effectively. Optimization-based meta-learning algorithms, like Model-Agnostic Meta-Learning (MAML) \cite{finnModelAgnosticMetaLearningFast2017} and Reptile \cite{nicholFirstOrderMetaLearningAlgorithms2018}, are known for their good generalization and fast adaptation on new heterogeneous tasks. They are also model-agnostic and can be applied to any gradient descent-based approaches, enabling applications in supervised learning and reinforcement learning.

In \cite{jiangImprovingFederatedLearning2019}, the authors drew parallels between meta-learning and FL. Meta-learning algorithms run in two phases: meta-training and meta-testing. The authors mapped the meta-training step in MAML to the FL global model training process, and the meta-testing step to the FL personalization process in which a few steps of gradient descent are performed on local data during local adaptation. They also show that FedAvg is analogous to the Reptile algorithm, and are in fact equivalent when all clients possess equal amounts of local data. Given the similarities in the formulations of meta-learning and FL algorithms, meta-learning techniques can be applied to improve the global FL model, while achieving fast personalization on the clients (Fig. \ref{data_model}d).

Per-FedAvg \cite{fallahPersonalizedFederatedLearning2020}, which is a variant of FedAvg built on top of the MAML formulation, has also been proposed. 
\begin{equation}
\label{eq:per-fedavg}
    \min_{w \in \mathbb{R}^{d}}F\left(w\right) := \frac{1}{C}\sum_{c=1}^{C}f_c\left(w-\alpha \nabla f_c\left(w\right)\right),
\end{equation}
where $\alpha>0$ is the step size. The cost function can be written as the average of meta-functions $F_1, \cdots, F_c$, where $F_c(w):= f_c(w-\alpha \nabla f_c(w))$ is the meta-function associated with client $c$. In contrast to the optimization objective of FedAvg in Eq. \eqref{eq:std_loss} which aims to learn a global model that performs well on most participating clients, the new goal is transformed to learn a good initial global model that performs well on a new heterogeneous task after it is updated with a few steps of gradient descent. This problem formulation is suitable for learning an improved global model initialization for stronger personalization on local data silos with heterogeneous distributions. 
However, this approach is computationally expensive due to the need to compute second-order gradients. To reduce computational overhead, the authors evaluated 2 forms of gradient approximations: (i) FO-MAML \cite{finnModelAgnosticMetaLearningFast2017}, which replaces the gradient estimate with its first-order approximation where the Hessian term is ignored; and (ii) HF-MAML \cite{fallahConvergenceTheoryGradientBased2020}, which replaces the Hessian-vector product with gradient differences. It has been found that HF-MAML achieves better gradient approximation.  

The idea of Per-FedAvg has been extended in \cite{dinhPersonalizedFederatedLearning2020} to propose a federated meta-learning formulation using Moreau envelopes (pFedMe). It incorporates an $l_2$-norm regularization loss which can control the balance between personalization and generalization performance. It has achieved improved accuracy and convergence over FedAvg and Per-FedAvg.

The authors in \cite{khodakAdaptiveGradientBasedMetaLearning2019} proposed the ARUBA framework. It is based on online learning to achieve adaptive meta-learning under FL settings. When combined with FedAvg, it improves model generalization performance and eliminates the need for hyperparameter optimization during personalization. \bigskip

\begin{table*}[t!] \small%
\caption{Summary of personalization techniques in \textit{Global Model Personalization}.}
\centering
\label{summ_1}
\resizebox{1\textwidth}{!}{
\begin{tabular}{|l|p{0.4\linewidth}|p{0.4\linewidth}|}
\hline
\textbf{Method} & \textbf{Advantages} & \textbf{Disadvantages}\\
\hline
Data Augmentation & 
\begin{itemize}[left=0pt,topsep=0pt]
\item Easy to implement, can be built on the general FL training procedure
\end{itemize}\nointerlineskip & \begin{itemize}[left=0pt,topsep=0pt]
\item Possibility of privacy leakage
\item May require a representative proxy dataset 
 \end{itemize}\nointerlineskip \\
\hline

Client Selection & 
\begin{itemize}[left=0pt,topsep=0pt]
\item Only modifies the client selection strategy of the general FL training procedure
\end{itemize}\nointerlineskip & \begin{itemize}[left=0pt,topsep=0pt]
\item Increased computational overhead from client subset optimization
\item May require a representative proxy dataset
\end{itemize}\nointerlineskip \\
\hline

Regularization & 
\begin{itemize}[left=0pt,topsep=0pt]
\item Easy to implement, slight modification to the FedAvg algorithm
\end{itemize}\nointerlineskip & \begin{itemize}[left=0pt,topsep=0pt]
\item Single global model setup
\end{itemize}\nointerlineskip \\
\hline

Meta-learning & 
\begin{itemize}[left=0pt,topsep=0pt]
\item Optimizes the global model for fast personalization
\end{itemize}\nointerlineskip & \begin{itemize}[left=0pt,topsep=0pt]
\item Single global model setup
\item Computationally expensive to compute second-order gradients
\end{itemize}\nointerlineskip \\
\hline

Transfer Learning & 
\begin{itemize}[left=0pt,topsep=0pt]
\item Improves personalization by reducing the domain discrepancy between the global and local models
\end{itemize}\nointerlineskip & \begin{itemize}[left=0pt,topsep=0pt]
\item Single global model setup
\end{itemize}\nointerlineskip \\

\hline
\end{tabular}
}
\end{table*}

\noindent\textit{Transfer Learning}
\newline
Transfer learning (TL) is commonly used for model personalization in non-federated settings \cite{panSurveyTransferLearning2010}. It aims to transfer knowledge from a source domain to a target domain, where both domains are often different but related. TL is an efficient approach that leverages knowledge transfer from a pre-trained model, thereby avoiding the need to build models from scratch. TL-based PFL approaches have also emerged. FedMD \cite{liFedMDHeterogenousFederated2019} is an FL framework based on TL and knowledge distillation for clients to design independent models using their own private data. Before the FL training and knowledge distillation phases, TL is first carried out using a model pre-trained on a public dataset. Each client then fine-tunes this model on its private data.

Domain adaptation TL techniques are commonly adopted to achieve PFL. These techniques aim to reduce the domain discrepancy between the trained global FL model (i.e., the source domain) and a given local model (i.e., the target domain) for improved personalization. There are several studies in FL that uses TL in the healthcare domain for model personalization (e.g., FedHealth \cite{chenFedHealthFederatedTransfer2019} and FedSteg \cite{yangFedStegFederatedTransfer2020}). The training procedure generally involves three steps: (i) training a global model via FL; (ii) training local models by adapting the global model on local data; and (iii) training personalized models by refining the local model using the global model via transfer learning. In order to enable domain adaptation, an alignment layer, such as the correlation alignment (CORAL) layer \cite{sunReturnFrustratinglyEasy2016}, is often added before the softmax layer for adaptation of the second-order statistics of the source and target domains (Fig. \ref{data_model}e).

To reduce training overhead in deep neural networks, the lower layers of the global model are often transferred and reused directly in the local models as low level generic features are learned. Other layers of the local model are fine-tuned with the local data to learn task-specific features for personalization. \bigskip

\noindent\textit{Summary}
\newline

In this section, we have discussed \textit{Data-based Approaches} and \textit{Model-based Approaches} for \textit{Global Model Personalization}. We now summarize and compare the personalization techniques in terms of their advantages and disadvantages (as shown in Table \ref{summ_1}).

Data-based approaches aim to reduce the statistical heterogeneity of client data distributions to tackle the problem of client drift. Data augmentation methods are easy to implement in the general FL training procedure. However, the applicability of these data augmentation methods is limited to some extent as the possibility of privacy leakage from data sharing has not been adequately addressed in existing designs. Data samples \cite{zhaoFederatedLearningNonIID2018a}, \cite{jeongCommunicationEfficientOnDeviceMachine2018b} or data statistics about the clients' data distributions \cite{duanSelfBalancingFederatedLearning2021} are often shared during the training process. Client selection methods improve the model generalization performance by optimizing the subset of participating clients for each FL communication round. As this requires computationally intensive algorithms such as deep Q-learning \cite{wangOptimizingFederatedLearning2020} and Multi-Armed Bandits \cite{yangFederatedLearningClass2020}, it incurs higher computational overhead than FedAvg. Additionally, many of these data-based approaches assume the availability of a proxy dataset that is representative of the global data distribution \cite{zhaoFederatedLearningNonIID2018a}, \cite{jeongCommunicationEfficientOnDeviceMachine2018b}, \cite{yangFederatedLearningClass2020}. In order to construct such a proxy dataset, it is necessary to understand the global data distribution, which is challenging under FL settings due to privacy-preservation concerns.

Model-based approaches closely follow the general FL training procedure in which a single global model is trained. Regularization methods such as FedProx \cite{liFederatedOptimizationHeterogeneous2020b} and MOON \cite{liModelContrastiveFederatedLearning2021} are easy to implement and they only require a slight modification to the FedAvg algorithm. Meta-learning optimizes the global model for fast personalization. However, gradient approximations are needed as it is expensive to compute second-order gradients \cite{finnModelAgnosticMetaLearningFast2017}, \cite{fallahPersonalizedFederatedLearning2020}. Transfer learning improves personalization by reducing the domain discrepancy between the global and local models. As the above approaches assume a single global model setup where a single global model is learnt over heterogeneous data silos, they are not well-suited for solution personalization when there are significant differences among the client data distributions. Additionally, model-based approaches generally assume that all clients and the FL server share a common model architecture. This assumption requires all clients to have sufficient computation and communication resources. However, edge computing FL clients are often resource-constrained \cite{lim2020federated}, making such approaches unsuitable.

\section{Strategy II: Learning Personalized Models}
\begin{figure*}[t!]
\centering
\includegraphics[width=1\linewidth]{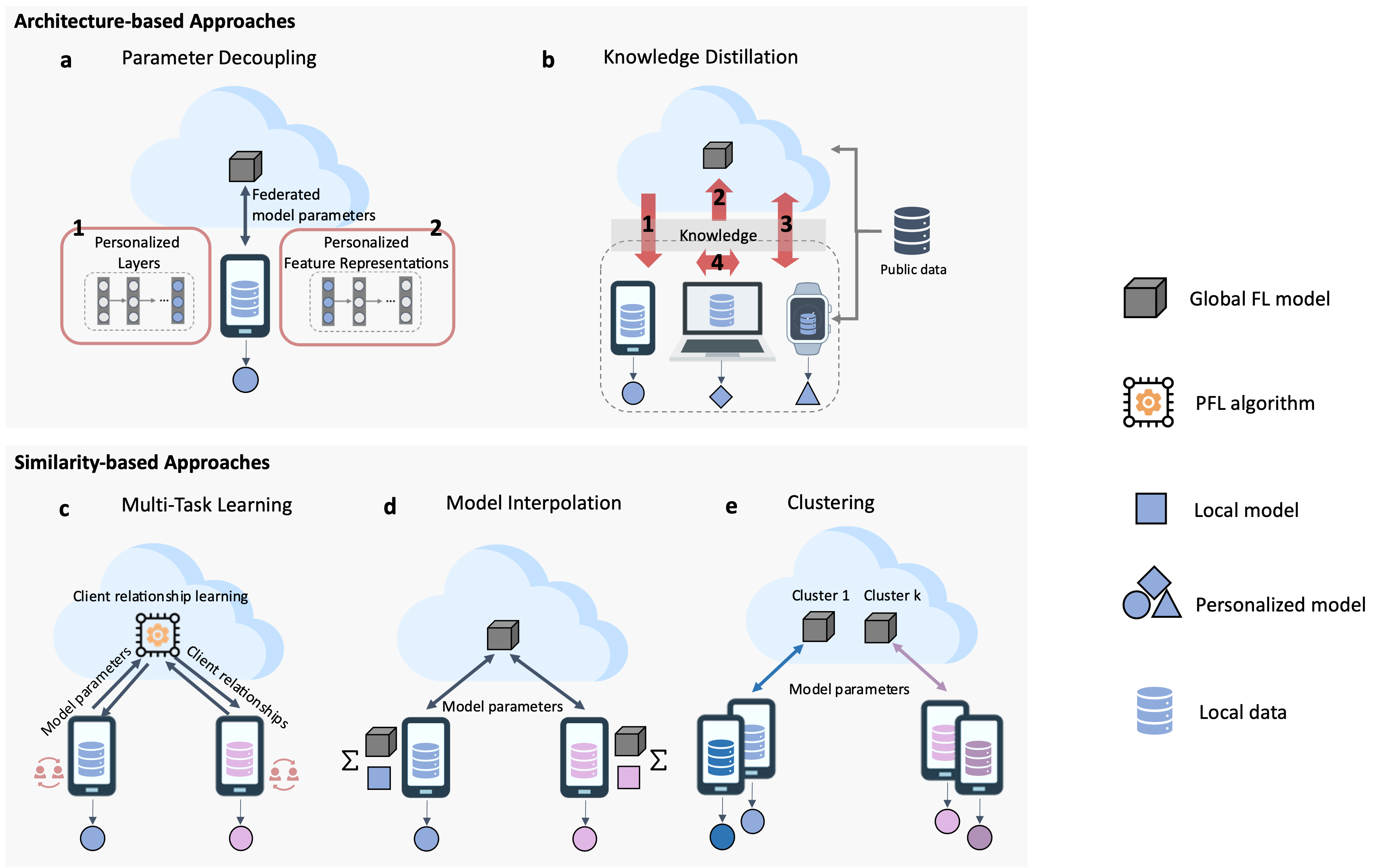}
\caption{The setup \& configurations of approaches that fall under \textit{Strategy II: Learning Personalized Models}. \textbf{a--b} Architecture-based approaches: (\textbf{a}) parameter decoupling; parameter privatization designs include \textbf{1}) personalized layers, \textbf{2}) personalized feature representations, (\textbf{b}) knowledge distillation; knowledge can be distilled \textbf{1}) towards clients, \textbf{2}) towards server, \textbf{3}) towards both clients \& server, \textbf{4}) amongst clients. \textbf{c--e} Similarity-based approaches: (\textbf{c}) multi-task learning, (\textbf{d}) model interpolation, (\textbf{e}) clustering.}
\label{arch_sim}
\end{figure*}

In this section, we survey PFL approaches following the strategy of learning personalized models. The main setup and configurations for these approaches are illustrated in Fig. \ref{arch_sim}. Based on our proposed taxonomy, they are divided  into \textit{Architecture-based Approaches} and \textit{Similarity-based Approaches} as follows.

\subsection{Architecture-based Approaches}
Architecture-based PFL approaches aim to achieve personalization through a customized model design that is tailored to each client. Parameter decoupling methods implement personalization layers for each client, while knowledge distillation methods support personalized model architectures for each client.
\newline

\noindent\textit{Parameter Decoupling}
\newline
Parameter decoupling aims to achieve PFL by decoupling the local private model parameters from the global FL model parameters. Private parameters are trained locally on the clients, and not shared with the FL server. This enables task specific representations to be learned for enhanced personalization.

The division between private and federated model parameters is an architectural design decision. There are generally two configurations used in parameter decoupling for deep feed-forward neural networks (Fig. \ref{arch_sim}a). The first is a ``base layers + personalized layers'' design proposed by \cite{arivazhaganFederatedLearningPersonalization2019}. In this setting, personalized deep layers are kept private by the clients for local training to learn personalized task-specific representations, while the base layers are shared with the FL server to learn low-level generic features.

The second design considers personalized feature representations for each client. In \cite{buiFederatedUserRepresentation2019}, a document classification model using a Bidirectional LSTM architecture is trained via FL by treating user embeddings as the private model parameters, and character embeddings (i.e., LSTM and MLP layers) as the FL model parameters. In \cite{liangThinkLocallyAct2020}, Local Global Federated Averaging (LG-FedAvg) has been proposed to combine local representation learning and global federated training. Learning lower dimensional local representations improves communication and computational efficiency for federated global model training. It also offers flexibility as specialized encoders can be designed based on the source data modality (e.g., images, texts). The authors also demonstrated how fair and unbiased representations that are invariant to protected attributes (e.g., race, gender) can be learned by incorporating adversarial learning into FL model training.

As the idea of parameter decoupling has some similarities to split learning (SL) \cite{guptaDistributedLearningDeep2018,vepakommaSplitLearningHealth2018}, a distributed and private machine learning paradigm, we briefly discuss their differences in this section. In SL, the deep network is split layer-wise between the server and the clients. Unlike parameter decoupling, the server model in SL is not transferred to the client for model training. Instead, only the weights of the split layer of the client model are shared during forward propagation and the gradients from the split layer are shared with the client during backpropagation. SL therefore has a privacy advantage over FL as the server and clients do not have full access to the global and local models \cite{thapaSplitFedWhenFederated2021}. However, training is less efficient due to the sequential client training process. SL also performs worse than FL on non-IID data and has higher communication overheads \cite{gaoEndtoEndEvaluationFederated2020}.

\bigskip
\noindent\textit{Knowledge Distillation}
\newline
In server-based horizontal federated learning (HFL) \cite{yangFederatedMachineLearning2019a}, the same model architecture is adopted by both the FL server and the FL clients. The underlying assumption is that there is sufficient communication bandwidth and computation capacity at the clients. However for practical applications with a large number of edge devices as FL clients, they are often resource-constrained. Clients may also choose to have different model architectures due to different training objectives. The key motivation for knowledge distillation in FL is to enable a greater degree of flexibility to accommodate personalized model architectures for the clients. At the same time, it also seeks to tackle communication and computation capacity challenges by reducing resource requirement.

Knowledge Distillation (KD) for neural networks was introduced by \cite{hintonDistillingKnowledgeNeural2015} as a paradigm for transferring the knowledge from an ensemble of teacher models to a lightweight student model. Knowledge is commonly represented as class scores or logit outputs in existing FL distillation approaches. In general, there are four main types for distillation-based FL architectures: (i) distillation of knowledge to each FL client to learn stronger personalized models, (ii) distillation of knowledge to the FL server to learn stronger server models, (iii) bi-directional distillation to both the FL clients and the FL server, and (iv) distillation amongst clients (Fig. \ref{arch_sim}b). 

In \cite{liFedMDHeterogenousFederated2019}, the authors proposed FedMD, a distillation-based FL framework which allows clients to design diverse models using their own private data via KD. Learning occurs through a consensus computed using the average class scores on a public dataset.  For every communication round, each client trains its model using the public dataset based on the updated consensus, and fine-tunes its model on its private dataset thereafter. This enables each client to obtain its own personalized model while leveraging knowledge from other clients. 
In \cite{zhuDataFreeKnowledgeDistillation2021}, the authors proposed FedGen, a data-free distillation framework that distills knowledge to the FL clients. A generative model is trained in the FL server and broadcast to the clients. Each client then generates augmented representations over the feature space using the learned knowledge as the inductive bias to regulate its local learning. 

In \cite{linEnsembleDistillationRobust2021}, the authors proposed the FedDF algorithm. It assumes a setting in which the edge clients require  different model architectures due to diverse computational capabilities. The FL server constructs $p$ distinct prototype models, each representing clients with identical model architectures (e.g., ResNet, MobileNet). For each communication round, FedAvg is first performed among clients from the same prototype group to initialize a student model. Cross-architecture learning is then performed via ensemble distillation, in which the client (teacher) model parameters are evaluated on an unlabelled public dataset to generate logit outputs that are used to train each student model in the FL server. 

Knowledge may also be distilled in a bi-directional manner between the FL client and FL server within the same FL training procedure. In \cite{annavaramGroupKnowledgeTransfer}, the authors proposed Group Knowledge Transfer (FedGKT) to improve model personalization performance for resource-constrained edge devices. It uses alternating minimization to train small edge models and a large server model through a bi-directional distillation approach. The large server model takes extracted features from the local models as inputs, and uses the  KL-divergence loss to minimize the difference between the ground truth and soft labels predicted by the local models. By doing so, the server model absorbs the knowledge transferred from the local models. Similarly, each local model calculates the KL-divergence loss using its private dataset and the predicted soft labels transferred from the server. This facilitates knowledge transfer from the server model to the local models. Using this bi-directional distillation framework, computation burden is shifted from the edge clients to the more powerful FL server. However, there is potential privacy risk as the ground truth labels from each client are uploaded to the FL server. 

KD-based PFL may also be carried out in distributed settings where knowledge is transferred amongst neighboring clients in a network. In \cite{bistritzDistributedDistillationOnDevice}, the authors proposed an architecture agnostic distributed algorithm for on-device learning -- D-Distillation. It assumes an IoT edge FL setting in which every edge device is connected to only a few neighboring devices. Only connected devices can communicate with each other. The learning algorithm is semi-supervised, with local training performed on private data and federated training on an unlabeled public dataset. For each communication round, each client broadcasts its soft decisions to its neighbors, while receiving their soft decision broadcasts. Each client then updates its soft decisions based on its neighbors' soft decisions via a consensus algorithm. The updated soft decisions are then used to update the client's model weights by regularizing its local loss. This procedure facilitates model learning via knowledge transfer amongst neighboring FL clients in a network.

\subsection{Similarity-based Approaches}
Similarity-based approaches aim to achieve personalization by modeling client relationships. A personalized model is learned for each client, with related clients learning similar models. Different types of client relationships have been studied in PFL. Multi-task learning and model interpolation consider pairwise client relationships, while clustering considers group level client relationships.
\newline

\noindent\textit{Multi-task Learning (MTL)}

The goal of multi-task learning (MTL) is to train a model that jointly performs several related tasks. This improves generalization by leveraging domain-specific knowledge across the learning tasks. By treating each FL client as a task in MTL, there is potential to learn and capture relationships among the clients exhibited by their heterogeneous local data (Fig. \ref{arch_sim}c). The MOCHA algorithm \cite{smithFederatedMultiTaskLearning2018} has been proposed to extend distributed MTL into the FL settings. MOCHA uses a primal-dual formulation to optimize the learned models. The algorithm addresses communication and system challenges prevalent in FL which are not considered in the field of MTL. Unlike the conventional FL design which learns a single global model, MOCHA learns a personalized model for each FL client. While MOCHA improves personalization, it is not suitable for cross-device FL applications as all clients are required to participate in every round of FL model training. Another drawback of MOCHA is that it is only applicable to convex models, and is thus unsuitable for deep learning implementations. This motivated \cite{corinziaVariationalFederatedMultiTask2019} to propose the VIRTUAL federated MTL algorithm that performs variational inference using a Bayesian approach. Although it can handle non-convex models, it is computationally expensive for large-scale FL networks.

In \cite{huangPersonalizedFederatedLearning2020}, the authors proposed FedAMP, an attention-based mechanism that enforces stronger pair-wise collaboration amongst FL clients with similar data distributions. In contrast to the standard FL framework in which a single global model is maintained by the server, FedAMP maintains a personalized cloud model $u_c$ for each client in the server. The personalized cloud model $u_c = \xi_{c,1}\theta_1+...+\xi_{c,m}\theta_m$ is the linear combination of the local client models $m \in C$, where $\sum_{m \in C}\xi_{c,m}=1$. In each communication round $t$, the personalized cloud model $u_c$ is transferred to client $c$ to perform local training on its own dataset. The local weights are computed as:
\begin{equation}
\theta_c^* = \argmin_{\theta \in \mathbb{R}^{d}}f_c\left(\theta\right)+\frac{\mu}{2\alpha}\lVert\theta-u_c\rVert^2
\end{equation} where $\alpha$ is the step size of gradient descent.

FedCurv \cite{shohamOvercomingForgettingFederated2019} uses EWC to prevent catastrophic forgetting when moving across learning tasks. Parameter importance is estimated using the Fisher information matrix and penalization steps are carried out to preserve important parameters. At the end of each communication round, each client sends its updated local parameters and the diagonal of its Fisher information matrix to the server. These parameters will be shared among all clients to perform local training in the next round. \bigskip

\noindent\textit{Model Interpolation}
\newline
In \cite{hanzelyFederatedLearningMixture2020}, a new formulation that learns personalized models using a mixture of global and local models has been proposed to balance generalization with personalization (Fig. \ref{arch_sim}d). Each FL client learns an individual local model. A penalty parameter $\lambda$ is used to discourage the local models from being too dissimilar from the mean model. Pure local model learning occurs when $\lambda$ is set to zero. This is equivalent to the fully personalized FL setting in Eq. \eqref{eq:full_per_loss} where each client trains its own model locally without any communication with other clients. As $\lambda$ increases, mixed model learning occurs and the local models become increasingly similar to each other. The setting approximates global model learning in which all local models are forced to be identical when $\lambda$ approaches infinity. In this way, the degree of personalization can be controlled. Additionally, the authors proposed a communication-efficient variant of SGD known as the Loopless Local Gradient Descent (L2GD). Through a probabilistic framework that determines whether a local GD step or a model aggregation step is to be performed, the number of communication rounds is reduced significantly.

In a related line of work, \cite{dengAdaptivePersonalizedFederated2020} proposed the APFL algorithm with the goal of finding the optimal combination of global and local models in a communication-efficient manner. They introduced a mixing parameter for each client which is adaptively learned during the FL training process to control the weights of the global and local models. This enables the optimal degree of personalization for each client to be learned. The weighting factor on a particular local model is expected to be larger if the local and global data distributions are not well-aligned, and vice versa. A similar formulation involving the joint optimization of local and global models to determine the optimal interpolation weight has been proposed in \cite{mansourThreeApproachesPersonalization2020a}.

Recently, \cite{diaoHeteroFLComputationCommunication2020} proposed the HeteroFL framework which trains local models with diverse computational complexities, based on a single global model. By adaptively allocating local models of different complexity levels according to the computation and communication capabilities of each client, it achieves PFL to address system heterogeneity in edge computing scenarios. \bigskip

\noindent\textit{Clustering}
\newline
For applications in which there are inherent partitions among clients or data distributions that are significantly different, adopting a client-server FL architecture to train a shared global model is not optimal. A multi-model approach in which an FL model is trained for each homogeneous group of clients is more suitable (Fig. \ref{arch_sim}e). Several recent works focus on clustering for FL personalization. The underlying assumption of clustering-based FL is the existence of a natural grouping of clients based on their local data distributions.

In \cite{sattlerClusteredFederatedLearning2019}, hierarchical clustering has been incorporated into FL as a post-processing step. An optimal bi-partitioning algorithm based on cosine similarity of the gradient updates from the clients is used divide the FL clients into clusters. As multiple communication rounds are needed to separate all incongruent clients, the recursive bi-partitioning clustering framework incurs high computation and communication costs that limit practical feasibility for large-scale settings. Another hierarchical clustering framework for FL has been proposed in \cite{briggsFederatedLearningHierarchical2020b}. It uses an agglomerative hierarchical clustering formulation that reduces clustering to a single step to lower computation and communication loads. The procedure begins by first training a global FL model for $t$ communication rounds. The global model is then fine-tuned on the private datasets of all clients to determine the difference $\Delta w$ between the global model parameters $w$ and the local model parameters $\theta_c$. The $\Delta w$ values for all clients are used as inputs to the agglomerative hierarchical clustering algorithm to generate multiple client clusters. FL training is then performed independently for each client cluster to produce multiple federated models. This approach is designed for a wider range of non-IID settings and allows training on a subset of clients during each round of FL model training. However, computing the pairwise distance between all clients in agglomerative clustering can be computationally intensive when there are a large number of clients. 

\begin{table*} \small%
\caption{Summary of personalization techniques in \textit{Learning Personalized Models}.}
\centering
\label{summ_2}
\resizebox{1\textwidth}{!}{
\begin{tabular}{|l|p{0.4\linewidth}|p{0.4\linewidth}|}
\hline
\textbf{Method} & \textbf{Advantages} & \textbf{Disadvantages}\\
\hline
Parameter Decoupling & 
\begin{itemize}[left=0pt,topsep=0pt]
\item Simple formulation
\item Layer-wise flexibility in architecture design for each client
\end{itemize}\nointerlineskip & \begin{itemize}[left=0pt,topsep=0pt]
\item Difficult to determine the optimal privatization strategy \end{itemize}\nointerlineskip \\
\hline

Knowledge Distillation & 
\begin{itemize}[left=0pt,topsep=0pt]
\item High degree of architecture design personalization for each client 
\item Communication-efficient
\item Supports resource heterogeneity \end{itemize}\nointerlineskip & \begin{itemize}[left=0pt,topsep=0pt]
\item Difficult to determine the optimal architecture design 
\item May require a representative proxy dataset 
\end{itemize}\nointerlineskip \\
\hline

Multi-Task Learning & 
\begin{itemize}[left=0pt,topsep=0pt]
\item Leverages pairwise client relationships to learn similar models for related clients
\end{itemize}\nointerlineskip & \begin{itemize}[left=0pt,topsep=0pt]
\item Sensitive to poor data quality of clients
\end{itemize}\nointerlineskip \\
\hline

Model Interpolation & 
\begin{itemize}[left=0pt,topsep=0pt]
\item Simple formulation using a mixture of global and local models
\end{itemize}\nointerlineskip & \begin{itemize}[left=0pt,topsep=0pt]
\item Uses a single global model as a basis for personalization 
\end{itemize}\nointerlineskip \\
\hline

Clustering & 
\begin{itemize}[left=0pt,topsep=0pt]
\item Good for applications where there are inherent partitions among clients
\end{itemize}\nointerlineskip & \begin{itemize}[left=0pt,topsep=0pt]
\item High computation and communication costs 
\item Additional system infrastructure for cluster management and deployment
\end{itemize}\nointerlineskip \\

\hline
\end{tabular}
}
\end{table*}

Other clustering approaches require a fixed number of clusters to be set at the beginning of FL training. In \cite{ghoshEfficientFrameworkClustered2020a}, the authors proposed the Iterative Federated Clustering Algorithm (IFCA). Instead of a single global model, the server constructs $K$ global models and broadcasts these models to all clients for local loss computation. Each client is assigned to one of the $K$ clusters the global model of which achieves the lowest loss value on the client's data. Cluster-based FL model aggregation within the cluster partition is then performed by the server. Compared to FedAvg, the communication overhead of IFCA is $K$ times higher as the server needs to broadcast $K$ cluster models to all clients in every communication round.

In \cite{huangPatientClusteringImproves2019}, the authors proposed community-based FL (CBFL) to predict patient hospitalization time and mortality. They trained a denoising autoencoder and performed K-means clustering with a pre-determined number of clusters to cluster patients based on the encoded features of their private data. An FL model is then trained for each cluster. 

In \cite{duanFedGroupEfficientClustered2021}, the authors proposed FedGroup, an FL clustering framework that implements a static client clustering strategy and a newcomer client cold start mechanism. FedGroup performs clustering on the local client updates using the K-means++ algorithm \cite{vassilvitskii2006k} based on the Euclidean distance of the Decomposed Cosine similarity (EDC).

In \cite{xieMultiCenterFederatedLearning2020}, the authors proposed a multi-center formulation that learns multiple global models. It introduces a new distance-based multi-center loss function:
\begin{equation}
\label{eq:multi_center_loss}
    \ell = \frac{1}{C}\sum_{k=1}^{K}\sum_{c=1}^{C}r_c^{(k)}Dist(\theta_c, w^{(k)}),
\end{equation} 
where $r_c^{(k)}$ means that client $c$ is assigned to cluster $k$, and $w^{(k)}$ is the model parameters of cluster $k$.
Expectation Maximization is used to solve the distance-based objective clustering problem and derive the optimal matching of clients to each cluster center. 
% The procedure is implemented in an iterative manner. 
In the E-step, the cluster assignment $r_c^{(k)}$ is updated by fixing $w_c$. $r_c^{(k)}$ is set to 1 if $k = \argmin_{j} Dist(\theta_c, w^{(j)})$. Otherwise, it is set to 0. In the M-step, the cluster centers $w^{(k)}$ are updated with
$w^{(k)} = \frac{1}{\sum_{c=1}^{C}r_c^{(k)}}\sum_{c=1}^{C}r_c^{(k)}w_c$. Finally, $w^{(k)}$ is sent to all clients in cluster $k$ to perform fine-tuning of the local model parameters $\theta_c$ on its private training data. The above steps are repeated until convergence.\bigskip

\noindent\textit{Summary}
\newline
In this section, we have discussed \textit{Architecture-based Approaches} and \textit{Similarity-based Approaches} for \textit{Learning Personalized Models}. We now summarize and compare the personalization techniques in terms of their advantages and disadvantages (as shown in Table \ref{summ_2}).

Architecture-based Approaches aim to achieve personalization through a customized model design that is tailored to each client. As parameter decoupling methods have a simple formulation that implement personalized layers for each client \cite{arivazhaganFederatedLearningPersonalization2019}, \cite{buiFederatedUserRepresentation2019}, it is limited in its ability to support a high degree of model design personalization. In contrast, KD-based PFL methods provide clients with a greater degree of flexibility to accommodate personalized model architectures for clients. They are also advantageous in communication and computation constrained edge FL settings \cite{linEnsembleDistillationRobust2021}, \cite{annavaramGroupKnowledgeTransfer}. However, a representative proxy dataset is often required in the KD process \cite{liFedMDHeterogenousFederated2019}, \cite{linEnsembleDistillationRobust2021}. For both methods, there are some challenges in model building. In parameter decoupling, the classification of private and federated parameters is an architectural design decision which controls the balance between generalization and personalization performance \cite{arivazhaganFederatedLearningPersonalization2019}. Determining the optimal privatization strategy is a research challenge. In KD, the effectiveness of knowledge transfer depends not only on model parameters, but also on model architecture. As it may be difficult for the student model to learn well if there is a huge capacity gap between the large teacher model and the small student model \cite{liuSearchDistillPearls,liBlockWiselySupervisedNeural2020}, it is imperative to determine an optimal design for both the server and client models.

Similarity-based approaches aim to achieve personalization by modeling client relationships. MTL methods such as FedAMP \cite{huangPersonalizedFederatedLearning2020} excel in capturing pairwise client relationships to learn similar models for related clients. As a result, it may be sensitive to poor data quality which result in the segregation of clients based on their data quality. Model interpolation methods have a simple formulation that learn personalized models using a mixture of global and local models. However, it is likely to experience a degradation in performance in highly non-IID scenarios as it uses a single global model as a basis for personalization \cite{dengAdaptivePersonalizedFederated2020}, \cite{diaoHeteroFLComputationCommunication2020}. Clustering methods are advantageous when there are inherent partitions among clients. However, they incur high computation and communication costs that limit practical feasibility for large-scale settings \cite{sattlerClusteredFederatedLearning2019}, \cite{briggsFederatedLearningHierarchical2020b}. Additional architectural components for the management and deployment of the clustering mechanism are also required \cite{briggsFederatedLearningHierarchical2020b}.

% **************************************************
% **************************************************
% **************************************************
% **************************************************
% **************************************************

\section{PFL Benchmark \& Evaluation Metrics} 
Another important factor for the long-term advancement of the PFL research field is performance benchmarking. In this section, we review and discuss the benchmarks and evaluation metrics used by existing PFL literature. \bigskip

\begin{table*}[htbp]
  \setlength{\aboverulesep}{0pt}
  \setlength{\belowrulesep}{0pt}
  \centering
  \caption{Types of non-IID data considered in PFL research.}
  \label{types-non-iid}
%  \resizebox{1\linewidth}{!}{
%   {\tabcolsep=2pt\def\arraystretch{1.1}
    \begin{tabular}{l|c|c|c|c}
    \toprule
    \textbf{Method} & \textbf{Quantity Skew} & \textbf{Feature Distribution Skew} & \textbf{Label Distribution Skew} & \textbf{Label Preference Skew} \\
    \midrule
    Data Augmentation & \cite{duanSelfBalancingFederatedLearning2021}     & \cite{wuFedHomeCloudEdgeBased2020}     & \cite{zhaoFederatedLearningNonIID2018a}--\cite{duanSelfBalancingFederatedLearning2021}     & - \\
    Client Selection & \cite{liFedSAENovelSelfAdaptive2021}     & \cite{chaiTiFLTierbasedFederated2020}--\cite{liFedSAENovelSelfAdaptive2021}     & \cite{wangOptimizingFederatedLearning2020}--\cite{liFedSAENovelSelfAdaptive2021}     & - \\
    \midrule
    Regularization & \cite{liFederatedOptimizationHeterogeneous2020b}     & \cite{liFederatedOptimizationHeterogeneous2020b}     & \cite{karimireddySCAFFOLDStochasticControlled}, \cite{liFederatedOptimizationHeterogeneous2020b}--\cite{liModelContrastiveFederatedLearning2021} & - \\
    Meta-Learning & \cite{dinhPersonalizedFederatedLearning2020}     & \cite{jiangImprovingFederatedLearning2019,khodakAdaptiveGradientBasedMetaLearning2019}     & \cite{fallahPersonalizedFederatedLearning2020,dinhPersonalizedFederatedLearning2020}     & - \\
    Transfer Learning & -    & \cite{liFedMDHeterogenousFederated2019}--\cite{yangFedStegFederatedTransfer2020}     & \cite{liFedMDHeterogenousFederated2019}     & - \\
    \midrule
    Parameter Decoupling & -      & \cite{arivazhaganFederatedLearningPersonalization2019}--\cite{buiFederatedUserRepresentation2019}     & \cite{arivazhaganFederatedLearningPersonalization2019,liangThinkLocallyAct2020}     & - \\
    Knowledge Distillation & -       & \cite{liFedMDHeterogenousFederated2019,zhuDataFreeKnowledgeDistillation2021,bistritzDistributedDistillationOnDevice}     & \cite{liFedMDHeterogenousFederated2019}, \cite{zhuDataFreeKnowledgeDistillation2021}--\cite{annavaramGroupKnowledgeTransfer}     & - \\
    \midrule
    Multi-Task Learning & -      & \cite{smithFederatedMultiTaskLearning2018}--\cite{huangPersonalizedFederatedLearning2020}     & \cite{huangPersonalizedFederatedLearning2020}--\cite{shohamOvercomingForgettingFederated2019}     & - \\
    Model Interpolation & -      & \cite{mansourThreeApproachesPersonalization2020a,hanzelyFederatedLearningMixture2020}     & \cite{hanzelyFederatedLearningMixture2020}--\cite{diaoHeteroFLComputationCommunication2020}     & - \\
    Clustering & \cite{xieMultiCenterFederatedLearning2020}     & \cite{sattlerClusteredFederatedLearning2019}--\cite{duanFedGroupEfficientClustered2021}, \cite{xieMultiCenterFederatedLearning2020}     & \cite{briggsFederatedLearningHierarchical2020b}, \cite{duanFedGroupEfficientClustered2021}     & \cite{sattlerClusteredFederatedLearning2019}--\cite{briggsFederatedLearningHierarchical2020b} \\
    \bottomrule
    \end{tabular}%
  \label{tab:addlabel}%
%  }
\end{table*}%
% }

\noindent\textit{FL Benchmark Datasets}
\newline
% Realistic datasets are important for the development of the PFL research field. 
There are several FL benchmarking frameworks developed in recent years, including FLBench \cite{liangFLBenchBenchmarkSuite2021}, Edge AIBench \cite{haoEdgeAIBenchComprehensive2018}, OARF \cite{huOARFBenchmarkSuite2020} and FedGraphNN \cite{he2021fedgraphnn}. LEAF \cite{caldasLEAFBenchmarkFederated2019} is one of the earliest and most popular benchmarking frameworks proposed for FL. At the time of writing, it provides six FL datasets covering a range of machine learning tasks including image classification, language modeling and sentiment analysis under both IID and non-IID settings. Examples datasets include the Extended MNIST \cite{cohen2017emnist} dataset split according to the writers of the character digits, the CelebA \cite{liu2015deep} dataset split according to the celebrity, and the Shakespeare \cite{mcmahanCommunicationEfficientLearningDeep2017} dataset split according to the characters in the play. A set of accuracy and communication metrics, along with implementation references for well-known approaches such as FedAvg, SGD and MOCHA are also provided. As LEAF extends existing public datasets from traditional machine learning settings, it does not fully reflect the data heterogeneity in FL scenarios. Although there are a few real-world federated datasets, such as a street image dataset for object detection \cite{luoRealWorldImageDatasets2019} and a species dataset for image classification \cite{hsuFederatedVisualClassification2020}, they are often limited in size. \bigskip

\noindent\textit{PFL Experimental Evaluation Design}
\newline
Despite the release of benchmark datasets for FL, they are not widely adopted in PFL research. The vast majority of PFL studies choose to simulate the non-IID setting by performing their own partitioning on a public benchmark dataset used in machine learning (e.g., MNIST \cite{lecun1998gradient}, EMNIST \cite{cohen2017emnist}, CIFAR-100 \cite{krizhevsky2009learning}), or creating a synthetic dataset \cite{mansourThreeApproachesPersonalization2020a,dengAdaptivePersonalizedFederated2020,ghoshEfficientFrameworkClustered2020a}. Here, we survey the different types of non-IID settings simulated in PFL literature and summarize them according to the personalization methods in Table \ref{types-non-iid}.

\subsubsection{Quantity Skew} FL clients hold local datasets of different sizes, with some clients having considerably larger amounts of data than others. Data size heterogeneity is pervasive in real-world environments due to diverse usage patterns across FL clients. To simulate data size heterogeneity, data from an imbalanced dataset are used directly without further sampling \cite{duanSelfBalancingFederatedLearning2021,xieMultiCenterFederatedLearning2020}. Alternatively, data can be distributed to FL clients according to power law \cite{liFedSAENovelSelfAdaptive2021,liFederatedOptimizationHeterogeneous2020b,dinhPersonalizedFederatedLearning2020}.

\subsubsection{Feature Distribution Skew} The feature distribution $P_c(x)$ varies across clients, while the conditional distribution $P(y|x)$ is the same across clients. For example, in health monitoring applications, the distributions of users' activity data vary considerably according to their habits and lifestyle patterns \cite{wuFedHomeCloudEdgeBased2020,chenFedHealthFederatedTransfer2019}. To model feature distribution skew, a dataset that is partitioned by users is often used with each user associated with a different client \cite{wuFedHomeCloudEdgeBased2020,smithFederatedMultiTaskLearning2018}. It can also be simulated by augmenting datasets via rotations \cite{ghoshEfficientFrameworkClustered2020a}.

\subsubsection{Label Distribution Skew} The label distribution $P_c(y)$ varies across clients, while the conditional distribution $P(x|y)$ is the same across clients. For example, in software mobile keyboards, label distribution skew is a likely problem for users from different demographics as there are diverse linguistic and cultural nuances that result in certain words or emojis to be used predominantly by different users. To model label distribution skew, the dataset is partitioned based on labels, where each client draws samples from a fixed number of label classes $k$. A smaller $k$ value would mean stronger data heterogeneity \cite{mcmahanCommunicationEfficientLearningDeep2017,chaiTiFLTierbasedFederated2020,liangThinkLocallyAct2020,dengAdaptivePersonalizedFederated2020}. Different levels of label distribution imbalance can be simulated by using a Dirichlet distribution $Dir(\alpha)$, where $\alpha$ controls the degree of data heterogeneity. An $\alpha$ of 100 is equivalent to the IID setting, while a smaller $\alpha$ value means that each client is more likely to hold data from only one class resulting in high data heterogeneity \cite{zhuDataFreeKnowledgeDistillation2021,linEnsembleDistillationRobust2021,hsuMeasuringEffectsNonIdentical2019}. 

\subsubsection{Label Preference Skew} The conditional distribution $P_c(x|y)$ varies across clients, while the label distribution $P(y)$ is the same across clients. Due to personal preferences, there may be variations in the labels. To model label preference skew, a proportion of labels are often swapped to increase variance in the ground truth labels \cite{sattlerClusteredFederatedLearning2019,briggsFederatedLearningHierarchical2020b}. 

From Table \ref{types-non-iid}, the evaluation of PFL algorithms is limited to a single type of non-IID setting in most existing studies. Feature distribution and label distribution skew are most commonly considered to simulate the non-IID setting in PFL studies. Label preference skew settings have only been adopted by clustering-based PFL approaches. Other PFL approaches have not been studied under this type of non-IID FL setting. A collective effort by the FL research community is needed to align and adopt benchmarks in order to standardization experimental evaluation design in PFL research.\bigskip

\begin{table*} %\small%
\caption{Evaluation metrics adopted by PFL research.}
\label{eval-metrics}
\centering
\resizebox{1\textwidth}{!}{
  \begin{tabular}{@{\extracolsep{3pt}}l>{\centering}p{0.13 \textwidth}>{\centering}p{0.13 \textwidth}>{\centering}p{0.17 \textwidth}>{\centering}p{0.08 \textwidth}>{\centering}p{0.1 \textwidth}>{\centering}p{0.08 \textwidth}>{\centering}p{0.07 \textwidth}>{\centering}p{0.05 \textwidth}>{\centering\arraybackslash}p{0.05 \textwidth}@{}}
%   \begin{tabular}{p{0.1 \linewidth}|l l|p{0.1 \linewidth}p{0.1 \linewidth}p{0.12 \linewidth}p{0.1 \linewidth}p{0.08 \linewidth}|ll}
    \toprule
    \multirow{2}{*}{\textbf{Method}} &
      \multicolumn{2}{c}{\textbf{Model Performance}} &
      \multicolumn{5}{c}{\textbf{System Performance}} &
      \multicolumn{2}{c}{\textbf{Trustworthy AI}} \\\cline{2-3} \cline{4-8} \cline{9-10}
      & {Accuracy} & {Convergence} & {Communication Efficiency} & {Computational Efficiency} & {System Heterogeneity} & {System Scalability} & {Fault Tolerance} & {Robustness} & {Fairness} \\
      \midrule
    Data Augmentation & \cite{zhaoFederatedLearningNonIID2018a}--\cite{wuFedHomeCloudEdgeBased2020} & \cite{duanSelfBalancingFederatedLearning2021}--\cite{wuFedHomeCloudEdgeBased2020} & \cite{jeongCommunicationEfficientOnDeviceMachine2018b}--\cite{wuFedHomeCloudEdgeBased2020} & - & - & \cite{duanSelfBalancingFederatedLearning2021} & - & - & - \\
    Client Selection & \cite{wangOptimizingFederatedLearning2020}--\cite{liFedSAENovelSelfAdaptive2021} & \cite{wangOptimizingFederatedLearning2020}--\cite{liFedSAENovelSelfAdaptive2021} & \cite{wangOptimizingFederatedLearning2020}--\cite{liFedSAENovelSelfAdaptive2021} & \cite{chaiTiFLTierbasedFederated2020} & \cite{chaiTiFLTierbasedFederated2020} & - & \cite{liFedSAENovelSelfAdaptive2021} & - & - \\ \hline
    Regularization & \cite{karimireddySCAFFOLDStochasticControlled}, \cite{liFederatedOptimizationHeterogeneous2020b}--\cite{liModelContrastiveFederatedLearning2021} & \cite{karimireddySCAFFOLDStochasticControlled}, \cite{liFederatedOptimizationHeterogeneous2020b}--\cite{liModelContrastiveFederatedLearning2021} & \cite{karimireddySCAFFOLDStochasticControlled}, \cite{yaoContinualLocalTraining2020}--\cite{liModelContrastiveFederatedLearning2021} & - & - & \cite{liFederatedOptimizationHeterogeneous2020b,liModelContrastiveFederatedLearning2021} & \cite{liFederatedOptimizationHeterogeneous2020b} & - & - \\
    Meta-Learning & \cite{jiangImprovingFederatedLearning2019}--\cite{fallahPersonalizedFederatedLearning2020}, \cite{dinhPersonalizedFederatedLearning2020}--\cite{khodakAdaptiveGradientBasedMetaLearning2019} & \cite{jiangImprovingFederatedLearning2019}--\cite{fallahPersonalizedFederatedLearning2020}, \cite{dinhPersonalizedFederatedLearning2020}--\cite{khodakAdaptiveGradientBasedMetaLearning2019} & \cite{jiangImprovingFederatedLearning2019,dinhPersonalizedFederatedLearning2020} & - & - & - & - & - & - \\
    Transfer Learning & \cite{liFedMDHeterogenousFederated2019}--\cite{yangFedStegFederatedTransfer2020} & \cite{liFedMDHeterogenousFederated2019} & \cite{liFedMDHeterogenousFederated2019} & - & - & - & - & - & - \\ \hline
    Parameter Decoupling & \cite{arivazhaganFederatedLearningPersonalization2019}--\cite{liangThinkLocallyAct2020} & \cite{arivazhaganFederatedLearningPersonalization2019}--\cite{liangThinkLocallyAct2020} & \cite{arivazhaganFederatedLearningPersonalization2019}--\cite{liangThinkLocallyAct2020} & - & - & - & - & \cite{liangThinkLocallyAct2020} & \cite{liangThinkLocallyAct2020} \\
    Knowledge Distillation & \cite{liFedMDHeterogenousFederated2019}, \cite{zhuDataFreeKnowledgeDistillation2021}--\cite{bistritzDistributedDistillationOnDevice} & \cite{liFedMDHeterogenousFederated2019}, \cite{zhuDataFreeKnowledgeDistillation2021}--\cite{linEnsembleDistillationRobust2021}, \cite{bistritzDistributedDistillationOnDevice} & \cite{liFedMDHeterogenousFederated2019}, \cite{zhuDataFreeKnowledgeDistillation2021}--\cite{linEnsembleDistillationRobust2021}, \cite{bistritzDistributedDistillationOnDevice} & \cite{annavaramGroupKnowledgeTransfer} & \cite{linEnsembleDistillationRobust2021} & - & \cite{zhuDataFreeKnowledgeDistillation2021} & - & - \\ \hline
    Multi-Task Learning & \cite{smithFederatedMultiTaskLearning2018}--\cite{shohamOvercomingForgettingFederated2019} & \cite{smithFederatedMultiTaskLearning2018}, \cite{huangPersonalizedFederatedLearning2020}--\cite{shohamOvercomingForgettingFederated2019} & \cite{huangPersonalizedFederatedLearning2020}--\cite{shohamOvercomingForgettingFederated2019} & - & \cite{smithFederatedMultiTaskLearning2018,huangPersonalizedFederatedLearning2020} & - & \cite{smithFederatedMultiTaskLearning2018,huangPersonalizedFederatedLearning2020} & - & - \\
    Model Interpolation & \cite{mansourThreeApproachesPersonalization2020a}, \cite{dengAdaptivePersonalizedFederated2020}--\cite{diaoHeteroFLComputationCommunication2020} & \cite{mansourThreeApproachesPersonalization2020a}, \cite{hanzelyFederatedLearningMixture2020}--\cite{diaoHeteroFLComputationCommunication2020} & \cite{hanzelyFederatedLearningMixture2020}--\cite{diaoHeteroFLComputationCommunication2020} & \cite{diaoHeteroFLComputationCommunication2020} & \cite{diaoHeteroFLComputationCommunication2020} & \cite{diaoHeteroFLComputationCommunication2020} & - & - & - \\
    Clustering & \cite{sattlerClusteredFederatedLearning2019}--\cite{duanFedGroupEfficientClustered2021}, \cite{xieMultiCenterFederatedLearning2020} & \cite{sattlerClusteredFederatedLearning2019}--\cite{duanFedGroupEfficientClustered2021}, \cite{xieMultiCenterFederatedLearning2020} & \cite{sattlerClusteredFederatedLearning2019}--\cite{briggsFederatedLearningHierarchical2020b}, \cite{huangPatientClusteringImproves2019}--\cite{duanFedGroupEfficientClustered2021}, \cite{xieMultiCenterFederatedLearning2020} & - & - & - & - & - & - \\
    \bottomrule
  \end{tabular}
  }
\end{table*}

\noindent\textit{PFL Evaluation Metrics}
\newline
We categorize the evaluation metrics adopted in PFL research into: 1) model performance related, 2) system performance related, and 3) trustworthy AI related (Table \ref{eval-metrics}). 

Model performance can be measured in terms of accuracy and convergence. Most PFL works adopt the average test accuracy of personalized models to measure model accuracy. While using an aggregated accuracy metric may be adequate to evaluate the performance of vanilla FL which trains a single globally-shared model, such a metric cannot reflect the performance of individual personalized models. As such, there are PFL works that use distribution-based evaluation frameworks such as histogram profiling \cite{huangPersonalizedFederatedLearning2020,wangFederatedEvaluationOndevice2019}, variance metrics \cite{fallahPersonalizedFederatedLearning2020,zhuDataFreeKnowledgeDistillation2021,liDittoFairRobust2021} and metrics at the individual client level \cite{wuFedHomeCloudEdgeBased2020,chenFedHealthFederatedTransfer2019} to evaluate the performance of personalized models. 
As each client experiences a different baseline accuracy due to statistical data heterogeneity, measuring the changes in model accuracy before and after personalization is a useful approach to assess the benefits of personalization \cite{yaoContinualLocalTraining2020,wangFederatedEvaluationOndevice2019,diviNewMetricsEvaluate2021}. Model convergence is measured by training loss \cite{liFedSAENovelSelfAdaptive2021,liModelContrastiveFederatedLearning2021,dengAdaptivePersonalizedFederated2020,sattlerClusteredFederatedLearning2019,briggsFederatedLearningHierarchical2020b}, number of communication rounds \cite{karimireddySCAFFOLDStochasticControlled,liModelContrastiveFederatedLearning2021}, number of local training epochs \cite{duanSelfBalancingFederatedLearning2021,liModelContrastiveFederatedLearning2021,dinhPersonalizedFederatedLearning2020}, and formalization of convergence bounds \cite{karimireddySCAFFOLDStochasticControlled,liFederatedOptimizationHeterogeneous2020b,fallahPersonalizedFederatedLearning2020}.

System performance metrics focus on communication efficiency, computational efficiency, system heterogeneity, system scalability and fault tolerance. Communication efficiency is evaluated by the number of communication rounds \cite{karimireddySCAFFOLDStochasticControlled,liModelContrastiveFederatedLearning2021}, the number of parameters \cite{duanSelfBalancingFederatedLearning2021,annavaramGroupKnowledgeTransfer,diaoHeteroFLComputationCommunication2020} and message sizes \cite{bistritzDistributedDistillationOnDevice,suiFedEDFederatedLearning2020}. Computational efficiency is evaluated in terms of the number of FLOPs \cite{annavaramGroupKnowledgeTransfer,diaoHeteroFLComputationCommunication2020} and training time \cite{chaiTiFLTierbasedFederated2020,annavaramGroupKnowledgeTransfer}. System heterogeneity is assessed by simulating variations in hardware capabilities and network conditions. This can be achieved by varying the number of local training epochs \cite{smithFederatedMultiTaskLearning2018,huangPersonalizedFederatedLearning2020}, CPU resources \cite{chaiTiFLTierbasedFederated2020} and local model complexity \cite{linEnsembleDistillationRobust2021,diaoHeteroFLComputationCommunication2020}. System scalability is evaluated in terms of performance on a large number of clients \cite{liModelContrastiveFederatedLearning2021}, total elapse time \cite{duanSelfBalancingFederatedLearning2021,liFederatedOptimizationHeterogeneous2020b} and total memory consumption \cite{liFederatedOptimizationHeterogeneous2020b,diaoHeteroFLComputationCommunication2020}. Fault tolerance is measured in terms of performance under different ratios of dropped out clients \cite{smithFederatedMultiTaskLearning2018,huangPersonalizedFederatedLearning2020} and stragglers \cite{liFederatedOptimizationHeterogeneous2020b,zhuDataFreeKnowledgeDistillation2021}. 

Trustworthy AI metrics have not been extensively adopted to evaluate PFL approaches. There are a few emerging works that consider these metrics \cite{diviNewMetricsEvaluate2021}. In \cite{liangThinkLocallyAct2020}, local model fairness and robustness against adversary attacks have been used to evaluate the performance of the proposed approach.  

The current direction for the evaluation of personalization performance in PFL research focuses primarily on accuracy gains in terms of model performance. However, the costs for achieving PFL should also be considered. While seeking accurate models, there are often trade-offs in terms of system scalability, as well as communication and computation overheads. The fulfillment of trustworthy AI attributes is also not sufficiently considered. It is important to design an effective PFL framework that jointly optimizes these cost-benefit objectives that are important in real-world FL applications. Given that PFL faces unique challenges and application scenarios, it is imperative to strengthen the development of evaluation metrics that are tailored to PFL.

\bigskip

\section{Promising Future Research Directions}
The field of PFL is starting to gain traction as practical FL applications begin to demand for models with better personalization performance. Based on our review of existing PFL literature, we envision promising future trajectories of research towards new PFL architectural design, realistic benchmarking, and trustworthy PFL approaches.

\subsection{Opportunities for PFL Architectural Design}
\textbf{Client Data Heterogeneity Analytics:} The heterogeneity of data among FL clients is a key consideration when assessing the type of PFL required. For example, a multi-model approach such as clustering is preferred for applications where there are inherent partitions or data distributions that are significantly different. In order to facilitate experimentation on non-IID data, recent works in PFL have proposed metrics like Total Variation, 1-Wasserstein \cite{fallahPersonalizedFederatedLearning2020} and Earth Mover’s Distance (EMD) \cite{zhaoFederatedLearningNonIID2018a} to quantify the statistical heterogeneity of data distributions. However, these metrics can only be calculated with access to raw data. The problem of FL client data heterogeneity analysis in a privacy-preserving manner remains open.

\textbf{Aggregation Procedure:} In more complex PFL scenarios, averaging-based model aggregation may not be an ideal approach in handling data heterogeneity. Model averaging is adopted in most prevailing FL architectures, and its effectiveness as an aggregation method has not been well-studied for PFL from a theoretical perspective \cite{xiaoAveragingProbablyNot2020}. Recently, \cite{wangFederatedLearningMatched2020} proposed a layer-wise matched averaging formulation for CNN and LSTM architectures. Specialized aggregation procedures for PFL are to be explored.

\textbf{PFL Architecture Search:} In the presence of statistical heterogeneity, federated neural architectures are highly sensitive to hyperparameter choices and may therefore experience poor learning performance if not tuned carefully \cite{liConvergenceFedAvgNonIID2020}. The choice of the FL model architecture also need to fit the underlying non-IID distribution well. Neural Architecture Search (NAS) \cite{zhuFederatedLearningFederated2021a} is a promising technique to help PFL reduce manual design effort to optimize the model architecture based on given scenarios. It will be particularly beneficial for parameter decoupling and knowledge distillation-based PFL methods.

%\subsection{Deploying PFL Approaches}
\textbf{Spatial Adaptability:} It refers to the ability of PFL systems to handle variations across client datasets as a result of (i) the addition of new clients, and/or (ii) dropouts and stragglers. These are practical issues prevalent in complex edge computing-based FL environments, where there is significant variability in hardware capabilities in terms of computation, memory, power and network connectivity \cite{wuPersonalizedFederatedLearning2020}.
%When a new client joins FL, the cold-start problem may occur. 

(i) Existing PFL approaches commonly assume a fixed client pool at the start of an FL training cycle, and that new clients cannot join the training process midway \cite{jeongCommunicationEfficientOnDeviceMachine2018b,briggsFederatedLearningHierarchical2020b}. Other approaches involve a pre-training step \cite{liFedMDHeterogenousFederated2019} that require time for local computation. Besides meta-learning approaches \cite{fallahPersonalizedFederatedLearning2020} that encourage fast learning on a new client, there is limited work addressing the cold-start problem in PFL. Current deep FL techniques are also prone to catastrophic forgetting of previously learned knowledge when new clients join, due to the stability-plasticity dilemma in neural networks \cite{kemkerMeasuringCatastrophicForgetting}. As a result, existing clients may experience a degradation in performance. A promising direction is to incorporate continual learning \cite{delangeContinualLearningSurvey2021} into FL to mitigate catastrophic forgetting.

(ii) With the prevalence of dropouts and stragglers in large-scale federated systems due to network, communication and computation constraints, it is necessary to design for robustness in FL systems. Developing communication-efficient algorithms to mitigate the problem of stragglers is an ongoing research direction, where gradient compression \cite{linDeepGradientCompression2020} and asynchronous model updates \cite{chenCommunicationEfficientFederatedDeep2020} are common strategies for addressing FL communication bottlenecks. These issues require further study in PFL to formalize the trade-offs between overhead and performance.

\textbf{Temporal Adaptability:} It refers to the ability of a PFL system to learn from non-stationary data. In dynamic real-world systems, we may expect changes in the underlying data distributions over time. This phenomenon is known as concept drift. Learning in the presence of concept drift often involve three steps: (i) drift detection (whether drift has occurred); (ii) drift understanding (when, how and where the drift occurs); and (iii) drift adaptation (response to drift) \cite{luLearningConceptDrift2018}. Casado et al. \cite{casadoConceptDriftDetection2020} is one of the few works that study the problem of concept drift in FL. It extends FedAvg with the Change Detection Technique (CDT) for drift detection. It remains an open direction to leverage existing drift detection and adaptation algorithms to improve learning on dynamic real-world data in PFL systems.

\subsection{Opportunities for PFL Benchmarking}  
\textbf{Realistic datasets:}
Realistic datasets are important for the development of a field. To facilitate PFL research, datasets that include more modalities like audio, video and sensor signals, and involve a broader range of machine learning tasks from real-world applications are required.

\textbf{Realistic non-IID settings:} In most existing studies, the evaluation of PFL algorithms is limited to a single type of non-IID setting. Experiments are performed by either leveraging an existing pre-partitioned public dataset (e.g., LEAF) or prepared by partitioning a public dataset to fit the target non-IID setting. For fairer comparison, it is imperative for the research community to develop a deeper understanding of the different non-IID settings in real-world federated learning in order to simulate realistic non-IID settings. Possible scenarios include: (i) temporal skew (changes in the data distributions over time) and (ii) the presence of adversarial attackers. Such an effort requires wider collaboration among researchers and industry practitioners, and will be beneficial for building a healthy PFL research ecosystem.

\textbf{Holistic evaluation metrics:} The establishment of systematic evaluation methodologies and metrics is important for PFL research. Model performance, system performance and Trustworthy AI attributes are important aspects to consider when evaluating the performance of an FL system. Methodologies that can provide a holistic cost-benefit analysis on a given PFL approach are needed for potential adopters to gain deeper insight into its real-world impact.

\subsection{Opportunities for Trustworthy PFL}  

\textbf{Open Collaboration:} Besides algorithmic challenges, future PFL research can explore promoting collaboration among self-interested data owners. For instance, data owners with personalized FL models may need to collaborate by sharing their models with other suitable data owners in order to adapt to changes in the learning task over time in dynamic real-world applications \cite{zheng2021incentive}. Incentive mechanism design is a promising research direction towards this vision. Game theory, pricing and auction mechanisms \cite{zhan2021survey} may be applied to build suitable incentive schemes to support the emergence of open collaborative PFL systems.

\textbf{Fairness:} As machine learning technologies become more widely adopted by businesses to support decision-making, there has been a growing interest in developing methods to ensure fairness in order to avoid undesirable ethical and social implications \cite{mehrabiSurveyBiasFairness2019,holsteinImprovingFairnessMachine2019}. 
Current approaches do not adequately address the unique set of fairness related challenges presented in PFL. These include new sources of bias introduced by the diversity of participating FL clients due to unequal local data sizes, activity patterns, location, and connection quality, etc. \cite{kairouzAdvancesOpenProblems2019a}. The study of fairness in PFL is still in its infancy and the framing of fairness in PFL has not yet been well-defined. The study of fairness in FL is mostly focused on the prevailing server-based FL paradigm \cite{mohriAgnosticFederatedLearning2019,liFairResourceAllocation2020,zhangHierarchicallyFairFederated2020}, although new work on fairness in alternative FL paradigms is emerging \cite{Lyu-et-al:2020TPDS}. As FL approaches maturity, advances in improving fairness for PFL in particular will become increasingly important in order for FL to be adopted at scale.

\textbf{Explainability:} Explainable Artificial Intelligence (XAI) \cite{XAI:2018} is an active research area that has attracted significant interest recently, driven by pressure from government agencies and the general public for interpretable models \cite{arrietaExplainableArtificialIntelligence2019}. It is important for models in high stake applications such as healthcare to be explainable, where there is a strong need to justify decisions made \cite{tonekaboniWhatCliniciansWant}.
Explainability has not yet been systematically explored in the FL literature. There are complex challenges unique to achieving explainability in PFL due to the scale and heterogeneity of distributed datasets. Striving for FL model explainability may also be associated with potential privacy risks from inadvertent data leakage, as demonstrated in \cite{shokriPrivacyRisksModel2021} where certain gradient-based explanation methods are prone to privacy leakage. There is few work addressing both explainability and privacy objectives simultaneously. Developing an FL framework that balances the trade-off between explainability and privacy is an important future research direction. One possible approach to achieve this trade-off is to incorporate explainability into the global FL model but not the personalization component of the FL model.

\textbf{Robustness:} Although FL offers better privacy protection compared to traditional centralized model training approaches, recent research has exposed vulnerabilities of FL that could potentially compromise data privacy \cite{lyuThreatsFederatedLearning2020a}. It is therefore of paramount importance to study FL attack methods and develop defensive strategies to counteract these attacks in order to ensure robustness of the FL system. With more complex protocols and architectures developed for PFL, more work is needed to study related forms of attacks and defenses to enable robust PFL approaches to emerge.

\section{Conclusions}
In this survey, we provide an overview of FL and discuss the key motivations for PFL. We propose a unique taxonomy of PFL techniques categorized according to the key challenges and personalization strategies in PFL, and highlight key ideas, challenges and opportunities for these PFL approaches. Finally, we discuss commonly adopted public datasets and evaluation metrics in PFL literature, and outline open problems and directions that would inspire further research in PFL. We believe that the discussions in this survey based on our proposed PFL taxonomy will serve as a useful roadmap for aspiring researchers and practitioners to enter the field of PFL and contribute to its long-term development.

\section*{Acknowledgments}
This research is supported, in part, by the National Research Foundation, Singapore under its AI Singapore Programme (AISG Award No: AISG2-RP-2020-019); Alibaba Group through Alibaba Innovative Research (AIR) Program and Alibaba-NTU Singapore Joint Research Institute (JRI) (Alibaba-NTU-AIR2019B1), Nanyang Technological University, Singapore; the RIE 2020 Advanced Manufacturing and Engineering (AME) Programmatic Fund (No. A20G8b0102), Singapore; the Nanyang Assistant Professorship (NAP); the Joint SDU-NTU Centre for Artificial Intelligence Research (C-FAIR) (NSC-2019-011); the NSFC No.91846205; National Key R\&D Program of China No. 2021YFF0900800; SDNSFC No. ZR2019LZH008; Shandong Provincial Key Research and Development Program (Major Scientific and Technological Innovation Project) (No. 2021CXGC010108); the National Key Research and Development Program of China under Grant 2018AAA0101100; and Hong Kong RGC TRS T41-603/20-R. Any opinions, findings and conclusions or recommendations expressed in this material are those of the authors and do not reflect the views of the funding agencies.

\bibliographystyle{IEEEtran}
\bibliography{main.bib}



% \end{thebibliography}

\vspace{-3em}
\begin{IEEEbiography}[{\includegraphics[width=1in,clip,keepaspectratio]{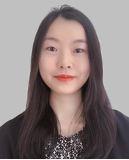}}]{Alysa Ziying Tan} is a PhD scholar at the Alibaba-NTU Joint Research Institute, Nanyang Technological University (NTU), Singapore. Her research interests include federated learning, deep learning and optimization. She obtained her Master's degree in Intelligent Systems from the National University of Singapore (NUS), and received the inaugural IMDA Singapore Digital Postgraduate Scholarship. Prior to her graduate studies, she worked as a data scientist and built deep learning and optimization solutions across manufacturing, insurance and supply chain domains.
\end{IEEEbiography}
\vspace{-2em}

\begin{IEEEbiography}[{\includegraphics[width=1in,clip,keepaspectratio]{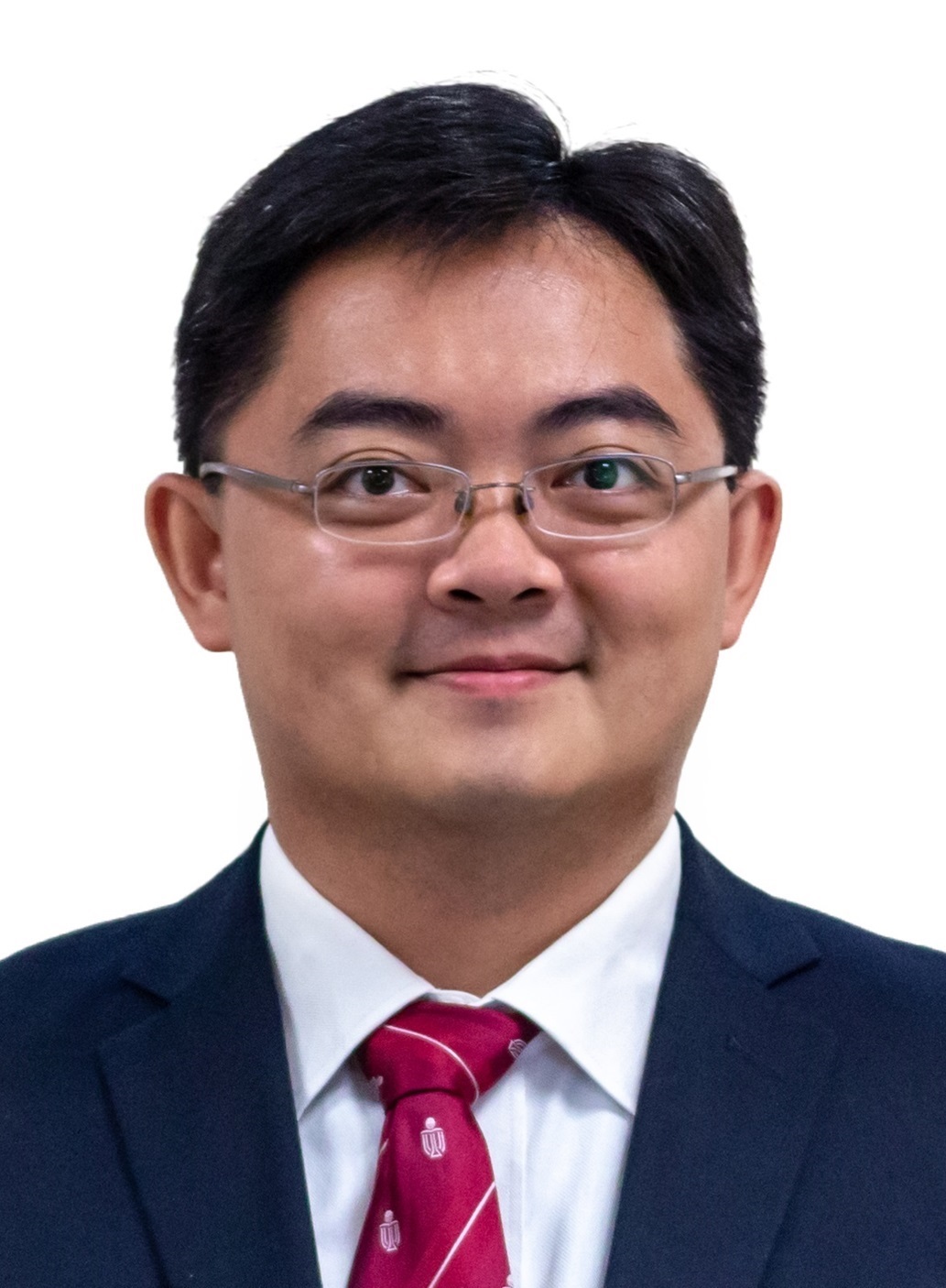}}]{Han Yu} is a Nanyang Assistant Professor (NAP) in the School of Computer Science and Engineering (SCSE), Nanyang Technological University (NTU), Singapore. He held the prestigious Lee Kuan Yew Post-Doctoral Fellowship (LKY PDF) from 2015 to 2018. He obtained his PhD from the School of Computer Science and Engineering, NTU. His research focuses on federated learning and algorithmic fairness. He has published over 150 research papers and book chapters in leading international conferences and journals. He is a co-author of the book \textit{Federated Learning} - the first monograph on the topic of federated learning. His research works have won multiple awards from conferences and journals.
\end{IEEEbiography}
\vspace{-2em}

\begin{IEEEbiography}[{\includegraphics[width=1in,clip,keepaspectratio]{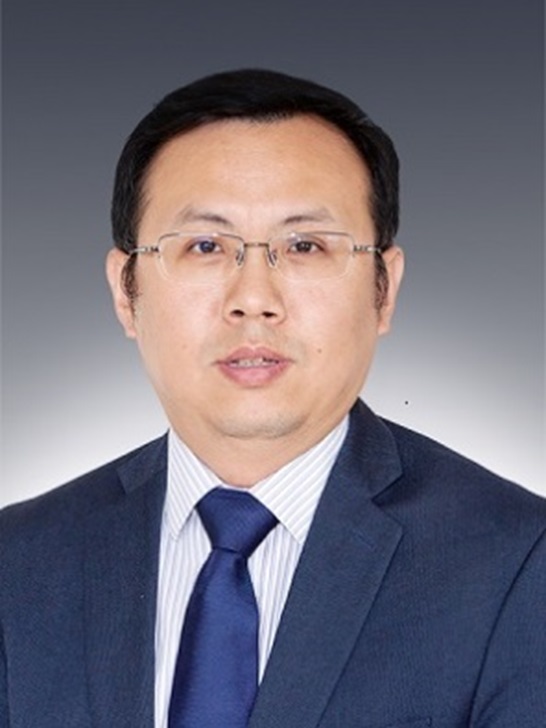}}]{Lizhen Cui} is a Professor and the Vice Chair of the School of Software Engineering, Shandong University. Between 2013 and 2014, he was a Visiting Scholar to the Georgia Institute of Technology in the United States. His main research interest includes data science and engineering, intelligent data analysis, service computing and collaborative computing.
\end{IEEEbiography}
\vspace{-2em}

\begin{IEEEbiography}[{\includegraphics[width=1in,clip,keepaspectratio]{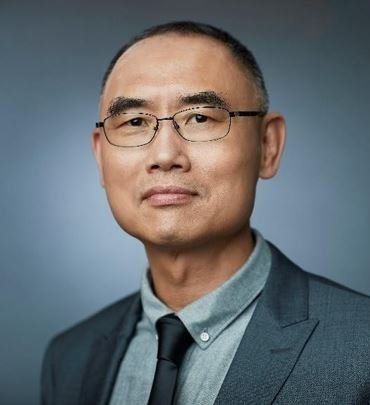}}]{Qiang Yang}
is the head of the AI Department at WeBank (Chief AI Officer) and Chair Professor at the Computer Science and Engineering (CSE) Department of the Hong Kong University of Science and Technology (HKUST), where he was a former head of CSE Department and founding director of the Big Data Institute (2015-2018). His research interests include AI, machine learning, and data mining, especially in transfer learning, automated planning, federated learning, and case-based reasoning. He is a fellow of several international societies, including ACM, AAAI, IEEE, IAPR, and AAAS. He received his Ph.D. in Computer Science in 1989 and his M.Sc. in Astrophysics in 1985, both from the University of Maryland, College Park. He obtained his B.Sc. in Astrophysics from Peking University in 1982. He had been a faculty member at the University of Waterloo (1989-1995) and Simon Fraser University (1995-2001). He was the founding Editor-in-Chief of the ACM Transactions on Intelligent Systems and Technology (ACM TIST) and IEEE Transactions on Big Data (IEEE TBD). He served as the President of International Joint Conference on AI (IJCAI, 2017-2019) and an executive council member of Association for the Advancement of AI (AAAI, 2016-2020). Qiang Yang is a recipient of several awards, including the 2004/2005 ACM KDDCUP Championship, the ACM SIGKDD Distinguished Service Award (2017), and AAAI Innovative AI Applications Award (2016). He was the founding director of Huawei's Noah's Ark Lab (2012-2014) and a co-founder of 4Paradigm Corp, an AI platform company. He is an author of several books including Intelligent Planning (Springer), Crafting Your Research Future (Morgan \& Claypool), and Constraint-based Design Recovery for Software Engineering (Springer).
\end{IEEEbiography}

\end{document}